\documentclass{article}

\usepackage[preprint]{neurips_2026}

\usepackage[utf8]{inputenc}
\usepackage[T1]{fontenc}
\usepackage{hyperref}
\usepackage{url}
\usepackage{booktabs}
\usepackage{amsfonts}
\usepackage{amsmath}
\usepackage{amssymb}
\usepackage{nicefrac}
\usepackage{microtype}
\usepackage{xcolor}
\usepackage{graphicx}
\usepackage{multirow}
\usepackage[font=small]{caption}

\title{Dynamics of the Transformer Residual Stream: Coupling Spectral Geometry to Network Topology}

\author{
  Jesseba Fernando \\
  Network Science Institute, Northeastern University \\
  \texttt{fernando.je@northeastern.edu}
  \And
  Grigori Guitchounts \\
  Flagship Pioneering \\
  \texttt{g.guitchounts@alumni.harvard.edu}
}

\begin{document}

\maketitle

\begin{abstract}
Large language models are remarkably capable, yet how computation propagates through their layers remains poorly understood. 
A growing line of work treats depth as discrete time and the residual stream as a dynamical system, where each layer's nonlinear update has a local linear description.
However, previous analyses have relied on scalar summaries or approximate linearizations, leaving the full spectral geometry of trained LLMs unknown.
We perform full Jacobian eigendecomposition across three production--scale LLMs and show that training installs a monotonic spectral gradient through depth---from non-normal, rotation-dominated early layers to near--symmetric late layers---together with a cumulative low-rank bottleneck that funnels perturbations into a small fraction of the residual stream's effective dimensions. 
Our experiments reveal that this gradient and the dimensional collapse are learned rather than architectural, and is largely dissolved when structured non-normality is removed. 
We further show that the topological positioning of graph communities predicts whether the Jacobian amplifies or suppresses them, with the sign of the coupling determined by the local operator type, a relationship absent at initialization.
These results map a learned spectral geometry in LLMs that links perturbation propagation and compression to the network's functional topology.
\end{abstract}

\section{Introduction}
\label{sec:introduction}

How computation propagates through a neural network remains largely unknown despite being one of the driving principles behind mechanistic interpretability. 
% MechInterp aims to understand how trained neural networks produce their outputs.
A growing toolkit addresses this question for transformers: circuit analysis traces computations through specific attention heads and MLP blocks \citep{olah2020zoom,olsson2022induction,lindsey2025biology}; sparse autoencoders decompose activations into interpretable features \citep{cunningham2023sparse,bricken2023monosemanticity}; probing classifiers and residual-stream readouts test what information is linearly accessible at each layer \citep{alain2016understanding,belrose2023tuned}.
These methods have been productive at identifying \emph{what} individual components represent.
But \emph{how} these components come to be, or how a perturbation at one layer becomes amplified, compressed or redirected as it passes through subsequent layers, is an open question. 

Dynamical systems theory is a natural formalism for this question, and viewing layers as discrete time steps has precedent in work that treats transformers as ODE solvers or interacting-particle systems \citep{lu_understanding_2019,geshkovski_mathematical_2024}, and in empirical analyses of residual-stream trajectories through depth \citep{hosseini_large_2023,lawson_residual_2024,fernando2025transformer}.
Each transformer block updates a shared residual stream, $\mathbf{h}_{\ell+1} = \mathbf{h}_\ell + f_\ell(\mathbf{h}_\ell)$, making depth a form of discrete time and the per-layer Jacobian $J_\ell = \partial \mathbf{h}_{\ell+1} / \partial \mathbf{h}_\ell$ the local linear description of the dynamics.
Theoretical literature on signal propagation has characterized these Jacobians at random initialization, establishing order-chaos phase boundaries \citep{poole2016exponential,schoenholz2017deep} and conditions for dynamical isometry \citep{pennington2017resurrecting}, but in most cases have been applied only at initialization or reduce the Jacobian to scalar summaries.
Recent empirical work has begun to probe Jacobian-like objects in \emph{trained} transformers---population-level linear maps \citep{fu2025cast}, detached Jacobians with frozen nonlinearities \citep{golden2025equivalent}, residual Jacobians correlated with benchmark performance \citep{aubry2025coupling}, Dynamic Mode Decomposition on averaged states \citep{jacobs2025block}---and each independently finds signatures of progressive dimensional funneling through depth.
These approaches use approximate or aggregate linearizations rather than the true input-space Jacobian. As a result, the full spectral geometry of trained production-scale Jacobians---the eigenvalue distribution, the non-normality structure, the balance of expanding and contracting modes, the rotational dynamics encoded in complex eigenvalues---remains uncharacterized.

A separate question concerns the relationship between a network's computational \emph{topology} and its \emph{dynamics}.
Prior work on modularity in neural networks has focused on the weight graph---showing that trained networks are more ``clusterable" than at initialization \citep{filan2021clusterability} or extracting communities from weight connectivity patterns \citep{watanabe2018community}.
In transformers, \citet{elhage2021framework} framed computation as independent components ``reading from'' and ``writing to'' a shared residual stream.
Whether this additive structure gives rise to mesoscale community organization in the residual stream's \emph{activation} correlations---and whether such structure is architectural or learned---has not been tested at production scale.

We address these gaps by computing exact Jacobians of three production-scale LLMs---Llama 3.1 8B, OLMo 3 7B, and Gemma 4 E4B---performing full eigendecomposition and singular value analysis at every layer.
For OLMo, we repeat this at three training checkpoints: step~0 (random initialization), step~471k, and step~1.41M (final pretraining step).
Because step~0 shares the trained model's architecture, depth, and wiring while varying only the weights, it serves as a null model: findings present at step~0 are architectural; findings absent at step~0 and emergent at trained checkpoints are attributed to training.
We bridge these dynamics to activation-graph topology by constructing activation-correlation graphs and running community detection to test agreement with the Jacobian dynamics across the three checkpoints noted above.

Our contributions are:

\begin{enumerate}
  \item \textbf{Full spectral characterization of trained production-scale Jacobians.}
  We find that per-layer Jacobians organize into depth regimes with a monotonic non-normality gradient---from near-rotational early layers to near-symmetric late layers.
  Approximately, 98\% of eigenvalues appear as complex conjugate pairs.
  Composing Jacobians across depth reveals that perturbations are funneled into a handful of effective channels ($\sim4-40$) across all three architectures.

  \item \textbf{Architecture-versus-training decomposition via OLMo's step-0 null model.}
  Depth regimes and adjacent-layer subspace decoupling are largely architectural, while the non-normality gradient and dimensional funneling are installed by training.
  Using Schur decomposition to zero the non-normality of the Jacobian reveals disruptions that track whether rank collapse is preserved, indicating that non-normal structure is central to the computations carried out by the residual stream.

  \item \textbf{Per-unit\footnote{We use ``unit'' throughout in the neuroscientific sense of a recording channel: the $i$-th coordinate of $\mathbf{h}_\ell \in \mathbb{R}^d$. 
  We avoid ``neuron'' because residual stream coordinates do not correspond to discrete computational elements but are bookkeeping channels of a shared workspace \citep{elhage2021framework}.} coupling between community topology and Jacobian dynamics.}
  Units that bridge multiple activation-graph communities (boundary nodes) are preferentially amplified or suppressed by the Jacobian, with the sign governed by operator type: boundary units are amplified at near-symmetric layers and de-amplified at non-normal ones.
  This coupling is absent at initialization and emerges monotonically over training ($0 \to 14 \to 24$ FDR-significant layers across OLMo checkpoints), establishing it as a learned property even though the mesoscale community structure itself is partly architectural.
\end{enumerate}

%%%%%%%%%%%%%%%%%%%%%%%%%%%%%%%%%%%%% RESULTS SECTION %%%%%%%%%%%%%%%%%%%%%%%%%%%%%%%%%%%%%
\section{Training reshapes per-layer Jacobians into a non-normality gradient and a dimensional bottleneck}
\label{sec:dynamics}

We treat each transformer block as one step of a discrete dynamical system on the residual stream and ask what its local linear description---the per-layer Jacobian $J_\ell = \partial \mathbf{h}_{\ell+1}/\partial \mathbf{h}_\ell$---looks like in trained transformers.
We compute exact per-sample Jacobians at every sub-layer boundary on 1{,}000 WikiText-2 \citep{wikitext} samples for three architectures (Llama 3.1 8B, OLMo 3 7B, Gemma 4 E4B) and, for OLMo, at three pre-training checkpoints (step~0, 471k, 1.41M). 
Step~0 shares the trained model's architecture, depth, and wiring while varying only the weights, so it serves as an architectural null. 
Computational and statistical details, and the formal definitions of the spectral quantities used below, are in Appendix~\ref{app:methods}.

\begin{figure}[!t]
\centering
\includegraphics[width=\linewidth]{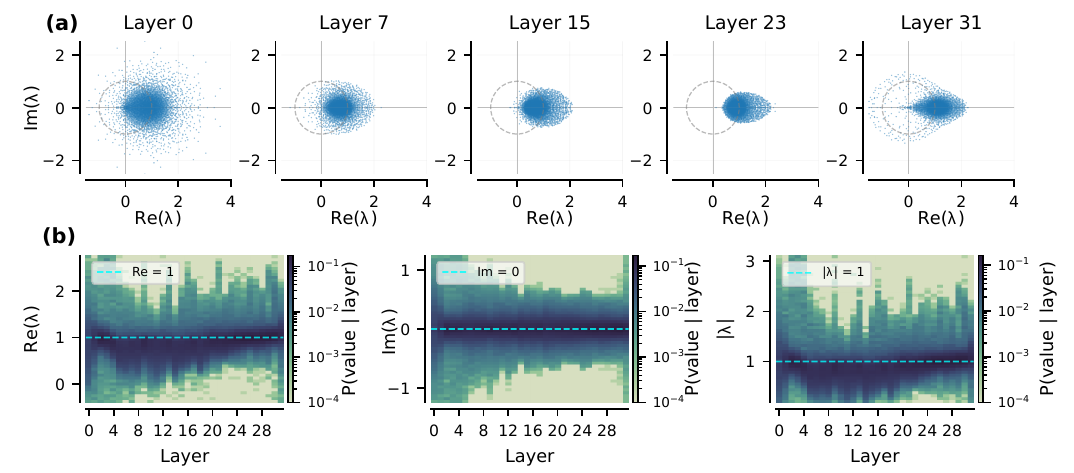}
\caption{Eigenvalue structure of Llama 3.1 8B mean Jacobians. \textbf{(a)}~Complex-plane scatter at five depths; the predominance of off-axis points reflects the $\sim$98\% complex conjugate pairs. \textbf{(b)}~Log-density heatmaps of $\mathrm{Re}(\lambda)$, $\mathrm{Im}(\lambda)$, and $|\lambda|$ across all 32 layers. Both panels show the three-regime transition: broad early, contracted mid, re-expanded late.}
\label{fig:eigenvalue-composite}
\end{figure}

A first basic observation cuts across every model and every layer: about 98\% of the eigenvalues of the mean Jacobian come as complex conjugate pairs (visualized for Llama in Figure~\ref{fig:eigenvalue-composite}; full per-layer densities for all three models in Figure~\ref{fig:eigenvalue-heatmaps-appendix}). 
Each such pair is a spiral---a two-dimensional subspace that is simultaneously rotated and either stretched or compressed---rather than a pure radial expansion or contraction. This rotational structure is invisible to the singular-value decompositions used in all prior spectral analyses of transformers \citep{fu2025cast, golden2025equivalent, aubry2025coupling}, which by construction collapse each mode to a non-negative scale factor and discard phase. Trained transformers thus route information through coupled rotational modes at every depth.

\subsection{Jacobians of trained transformers sweep from rotation-dominated to near-symmetric operators across depth}
\label{sec:gradient}

\begin{figure}[!t]
\centering
\includegraphics[width=\linewidth]{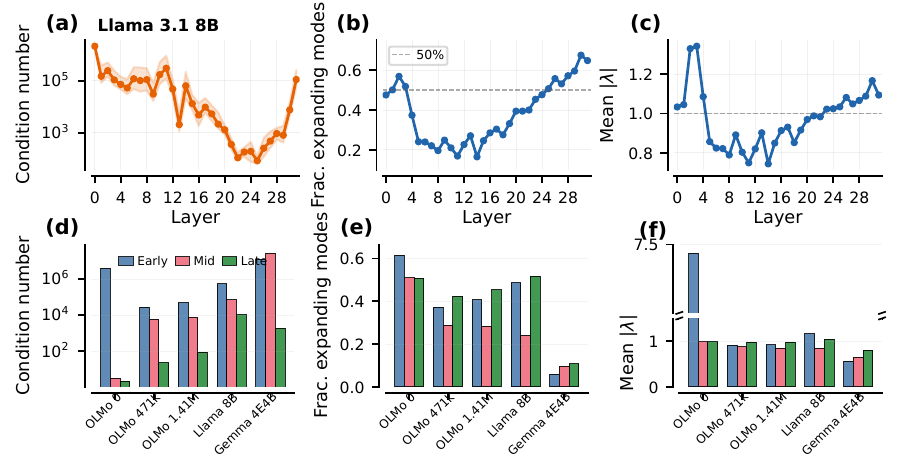}
\caption{Three-regime Jacobian structure across depth. \textbf{Top row:} Llama 3.1 8B per-layer profiles. \textbf{(a)}~Condition number $\kappa$ (median over 1{,}000 samples; shaded band shows IQR), \textbf{(b)}~fraction of expanding eigenvalue modes, \textbf{(c)}~mean eigenvalue magnitude. \textbf{Bottom row:} regime-mean summaries (early 0--4, mid 5--19, late 20--31) across all five configurations for the same three quantities: \textbf{(d)}~$\kappa$, \textbf{(e)}~fraction expanding, \textbf{(f)}~mean $|\lambda|$ (broken $y$-axis).}
\label{fig:section2-1}
\end{figure}

We set out to ask whether different layers play distinct dynamical roles, and if so, how those roles are distributed along the depth axis. 
Two complementary measures organize the layers (Figure~\ref{fig:section2-1}).
The condition number $\kappa_\ell = \sigma_{\max}/\sigma_{\min}$ summarizes geometric distortion; self-alignment $\|V_{\ell,:k}^\top U_{\ell,:k}\|_F^2 / k$ (overlap between leading input and output singular subspaces, $k = 64$) summarizes operator type, equaling 1 for symmetric operators and $k/d \approx 0.016$ for pure rotators.
In Llama 3.1 8B these measures partition the 32 layers into three regimes.
Early layers (0--4) are highly anisotropic ($\kappa \sim 10^6$) and strongly non-normal (self-alignment $\sim 0.04$): near-pure rotators.
Mid layers (5--19) collapse the condition number to $\sim 10^2$--$10^3$ and the expanding-mode fraction to 17--30\%, but remain non-normal.
Late layers (20--31) re-expand $\kappa$ to $\sim 10^5$--$10^6$ while rising toward symmetry (self-alignment $\sim 0.55$). Similar patterns were observed for OLMo and Gemma (Figure~\ref{fig:section2-1-appendix}).

\begin{figure}[!t]
\centering
\includegraphics[width=\linewidth]{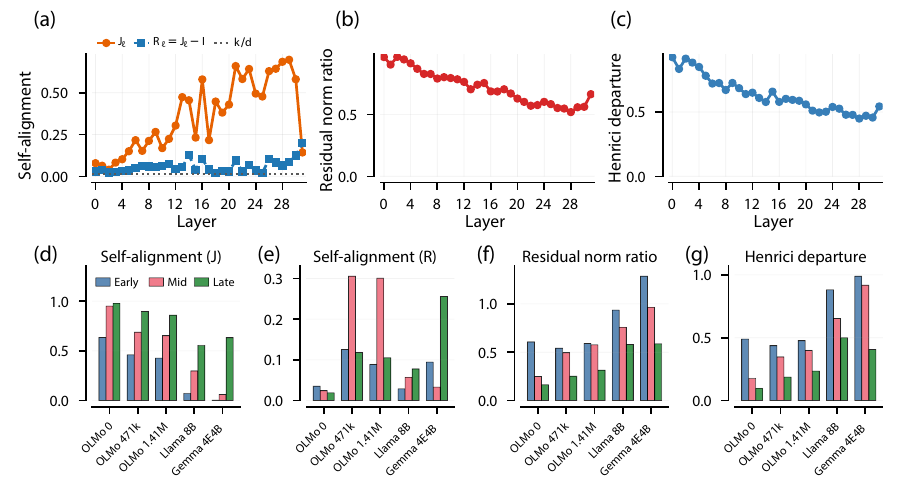}
\caption{Non-normality gradient across depth. \textbf{Top row:} Llama 3.1 8B per-layer profiles. \textbf{(a)}~Self-alignment of $J_\ell$ (orange) and $R_\ell = J_\ell - I$ (blue), with random baseline $k/d \approx 0.016$ (dotted). \textbf{(b)}~Residual norm ratio $\|R_\ell\|_F / \|J_\ell\|_F$. \textbf{(c)}~Henrici departure from normality $\delta(J_\ell)$. \textbf{Bottom row:} regime means (early/mid/late) across all five configurations for: \textbf{(d)}~$J$ self-alignment, \textbf{(e)}~$R$ self-alignment, \textbf{(f)}~residual norm ratio, \textbf{(g)}~Henrici departure. Full per-layer profiles for all models in Figure~\ref{fig:nonnormality-appendix}.}
\label{fig:non-normality-composite}
\end{figure}

Self-alignment rises monotonically from $\approx 0.04$ (layer~2) to $\approx 0.70$ (layer~29); the Henrici departure $\delta(J) = \sqrt{\|J\|_F^2 - \sum |\lambda_i|^2}\,/\,\|J\|_F$ falls in lockstep, from $0.91$ to $0.47$ (Figure~\ref{fig:non-normality-composite}a,c).
Operator type thus changes continuously from rotation toward symmetry even though geometric distortion bottoms out in the middle.
Could this rising symmetry merely reflect the identity in the skip connection dominating in late layers?
Stripping the skip ($R_\ell = J_\ell - I$) rules that out: $R_\ell$'s self-alignment stays below $\sim 0.20$ everywhere (Figure~\ref{fig:non-normality-composite}a).
The block computation is non-normal at every layer; the symmetry of $J_\ell$ instead tracks the residual norm ratio $\|R_\ell\|_F / \|J_\ell\|_F$, which declines from $0.963$ to $0.518$ as the identity progressively dominates (Figure~\ref{fig:non-normality-composite}b). The same pattern holds across all models (Figure~\ref{fig:nonnormality-appendix}).

OLMo's step-0 checkpoint separates architectural from learned contributions.
At initialization, all layers beyond the first are nearly symmetric (mid 0.95, late 0.98) with eigenvalue clouds collapsed near $\mathrm{Re}(\lambda) = 1$ (Figure~\ref{fig:2-1e}a; full trajectory in Figure~\ref{fig:olmo-eigenvalue-heatmaps-appendix}).
Training leaves the late ceiling near-symmetric ($0.98 \to 0.86$) while pulling early and mid layers toward rotation (mid $0.95 \to 0.65$; early $0.64 \to 0.43$; Table~\ref{tab:selfalign-regimes}).
The non-normality gradient is thus a joint product of architecture (late-regime ceiling) and training (early/mid rotator regime).

Does the gradient imply inter-layer coordination---do the leading singular directions one layer writes into match those the next layer reads?
Largely not: forward alignment $\|U_\ell^\top V_{\ell+1}\|_F^2 / k$ sits at or near the random baseline $k/d \approx 0.016$ across all models (Figure~\ref{fig:forward-alignment-appendix}; Appendix~\ref{app:forward-alignment}).
The residual stream thus functions as the shared workspace theorized by \citet{elhage2021framework}, with at most weak inter-layer coordination.

\subsection{Training installs a cumulative low-rank bottleneck through depth}
\label{sec:cumulative}

To capture end-to-end propagation we form the cumulative Jacobian product $P_\ell = J_{31} \cdots J_\ell$, mapping a perturbation at layer~$\ell$ to the final residual-stream state, and measure its dimensionality via effective rank $\mathrm{erank}(P_\ell) = \exp(-\sum_i p_i \log p_i)$ with $p_i = \sigma_i^2 / \sum_j \sigma_j^2$.

\begin{figure}[!t]
\centering
\includegraphics[width=\linewidth]{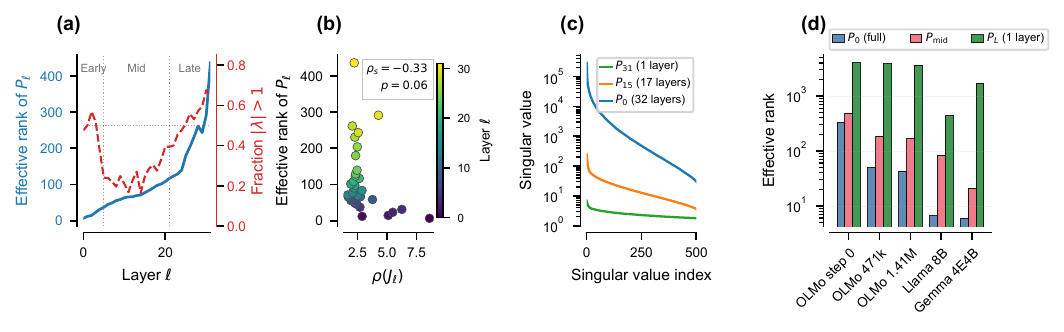}
\caption{Cumulative Jacobian analysis (Llama 3.1 8B unless noted). \textbf{(a)}~Effective rank of $P_\ell = J_{31} \cdots J_\ell$ (blue, left axis) and fraction of expanding eigenvalues (red dashed, right axis; dotted = 50\%) vs.\ injection layer. \textbf{(b)}~Effective rank vs.\ spectral radius $\rho(J_\ell)$, colored by layer; Spearman $\rho_s$ annotated. \textbf{(c)}~Singular value spectra of cumulative products at three depths. \textbf{(d)}~Effective rank of $P_0$ (full product), $P_{\text{mid}}$, and $P_L$ (single-layer) across all five configurations (broken $y$-axis).}
\label{fig:cumulative-jacobian}
\end{figure}

In Llama 3.1 8B, effective rank drops from $\sim 436$ (single-layer, $\ell = 31$) to $6.7$ at full composition ($\ell = 0$): of $4{,}096$ input directions, only about seven survive end-to-end (Figure~\ref{fig:cumulative-jacobian}a).
Gemma funnels to $5.9$ across 42 layers (Figure~\ref{fig:cumulative-jacobian-appendix}).
The collapse is not architectural: at OLMo step~0 effective rank stays high ($\approx 326$ for $P_0$, $\approx 4006$ for $P_{31}$); by step~1.41M it falls to $\sim 42$---two orders of magnitude below initialization (Figure~\ref{fig:cumulative-jacobian}d).
The steepest collapse occurs at early layers, where large spectral radii amplify a few eigendirections and suppress the rest (Spearman $\rho_s = -0.33$, $p = 0.06$; Figure~\ref{fig:cumulative-jacobian}b).

% The shape of the end-to-end singular spectrum makes the consequences concrete (Figure~\ref{fig:cumulative-jacobian}c). 
% $P_{31}$ (single layer) has a nearly flat spectrum with $\sigma$ in the range $1$--$7$, consistent with a mildly expanding operator. 
% Composing 17 layers ($P_{15}$) introduces two orders of magnitude of dynamic range ($\sigma_1 \approx 250$, $\sigma_{500} \approx 1$). 
% The full 32-layer product $P_0$ reaches $\sigma_1 \approx 3 \times 10^5$ while the tail collapses below 1: amplification is extreme but concentrated in a tiny subspace. 
% The trained network's end-to-end sensitivity is effectively low-rank, with the surviving channels experiencing $10^5$-fold gain and most input perturbations exponentially suppressed.

The trained network's sensitivity to its own input is thus confined to a subspace three orders of magnitude smaller than the residual stream---a learned bottleneck that bounds from above how many independent features any single forward pass can propagate from embedding to logit.

\subsection{The cumulative bottleneck is a property of the trained non-normal feedforward, not the spectrum}
\label{sec:schur_surgery}

The bottleneck in the residual stream could in principle reflect either of two distinct properties of $J_\ell$: the \emph{eigenvalue spectrum}, which we have shown contains expanding modes at every depth, or the trained \emph{non-normal feedforward}, the off-spectrum mass that distinguishes a generic operator from a normal one with the same eigenvalues. 
Applying Schur decomposition to the Jacobians allowed us to cleanly separate the two possibilities: $J_\ell = Q_\ell (\Lambda_\ell + N_\ell) Q_\ell^*$, with $\Lambda_\ell$ diagonal (the eigenvalues) and $N_\ell$ strictly upper-triangular (the non-normal piece). 
Holding $\Lambda_\ell$ and $Q_\ell$ fixed and scaling only $N_\ell$ by a dose $c \in \{0, 0.25, 0.5, 0.75, 1, 1.5, 2\}$ ($c=0$ removes all non-normality at fixed spectrum; $c=1$ is the trained model), we recomposed the linearized stack and recomputed $\mathrm{erank}(P_{0:31})$.

\begin{figure}[!h]
\centering
\includegraphics[width=\linewidth]{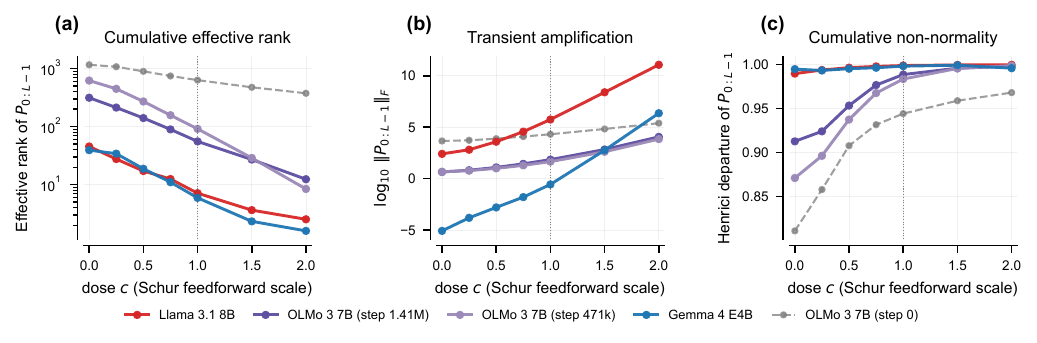}
\caption{Schur surgery on the trained non-normal feedforward of each layer's Jacobian. Each $J_\ell$ is written in complex Schur form and only $N_\ell$ is scaled by a dose $c$, holding the spectrum $\Lambda_\ell$ and basis $Q_\ell$ fixed; $c=0$ is fully normal at the trained spectrum, $c=1$ is the trained model. \textbf{(a)}~Cumulative effective rank of $P_{0:31}$ vs.\ dose (log $y$); all three trained models trace the same monotone curve, OLMo step~0 (untrained) is nearly flat. \textbf{(b)}~$\log_{10}\|P_{0:L}\|_F$ vs.\ dose: cumulative Jacobians gain several orders of magnitude in cumulative Frobenius norm as $N_\ell$ is scaled in, consistent with the transient-amplification reading of non-normality. \textbf{(c)}~Henrici departure of $P_{0:L}$ vs.\ dose: cumulative non-normality saturates near unity at modest $c$ in trained networks. Random-replacement controls and mode-$R$ confirmation in Appendix~\ref{app:schur-surgery}.}
\label{fig:schur-surgery}
\end{figure}

Removing the trained non-normal feedforward component of $J$, while keeping the eigenspectrum intact largely dissolves the bottleneck. 
In Llama 3.1 8B, $\mathrm{erank}(P_{0:31})$ rises from $7.1$ at $c=1$ to $45.4$ at $c=0$ (a $6.4\times$ recovery), and falls further to $2.5$ at $c=2$ (Figure~\ref{fig:schur-surgery}a); OLMo (step~1.41M) and Gemma 4 E4B reproduce the same monotone trajectory, with $c=0/c=1$ ratios of $5.7$ and $6.6$, while the untrained OLMo step~0 baseline is nearly flat over the same dose range. 
Similarly, we ruled out that the dimensionality collapse relies on the Jacobian's Frobenius-mass or the skip connection (Appendix~\ref{app:schur-surgery}, Figure~\ref{fig:schur-surgery-appendix})
The dimensional collapse in the transformer residual stream is therefore not an effect of simple properties of J or its eigenspectrum, but rather is a property of the trained upper-triangular structure. 

% Nor is it a Frobenius-mass effect: replacing $N_\ell$ at every layer with an i.i.d.\ Gaussian draw matched in $\|\cdot\|_F$ \emph{reverses} the dose response, raising $\mathrm{erank}(P_{0:31})$ to $216 \pm 1$ at $c=1$ in Llama (vs.\ $7.1$ for the trained $N_\ell$; Appendix~\ref{app:schur-surgery}, Figure~\ref{fig:schur-surgery-appendix}). 
% Applying the same surgery to the residual operator $R_\ell = J_\ell - I$ gives an identical pattern, ruling out an artifact of the $+I$ skip. 
% The cumulative bottleneck is mechanistically tied to the trained non-normal structure of $N_\ell$ in each layer---the same off-diagonal mass that governs per-layer self-alignment and Henrici departure in \S\ref{sec:gradient}.

% The operator-type gradient and the per-unit reading of Jacobian column norms as amplification both set up the next section, where we ask whether the residual stream's topological organization predicts which units that amplification falls on.

\section{Activation-graph community structure predicts which units the Jacobian amplifies, with a sign set by operator type}
\label{sec:topology}

Having characterized the low-dimensional, rotational dynamics of the transformer residual stream, we next asked if those dynamics are related to the stream's topological structure. 
Do the network's residual-stream \textit{units} cluster into functional groups?
And if so, where does community structure come from: is it a learned organization built by training, or is it already wired in by the architecture? Finally, does a unit's position in the community structure predict how the Jacobian treats it?

\subsection{Activation correlations form mesoscale communities at every depth}
\label{sec:topology-setup}

For each of the sub-layer (i.e. pre-Attention and pre-MLP, making 64 steps for a 32-layer model) snapshots we extract last-token activations on the same $1{,}000$ WikiText samples used in \S\ref{sec:dynamics}, build a sparse signed correlation graph by retaining the top-$k = 20$ edges per unit (positive and negative weights both kept), and partition the graph using signed Leiden CPM \citep{traag_narrow_2011, traag2019louvain} at resolution $\gamma = 0.001$ (see Appendix~\ref{app:methods} and Table~\ref{tab:community_method_comparison_overall} for joint $\gamma_{pos}/\gamma_{neg}$ formulation, justification and comparison methods)
This yielded $6$--$87$ non-degenerate communities per layer at every checkpoint of every model. 

% We chose $\gamma = 0.001$ on partition-stability grounds; three alternative methods ($|\text{corr}|$-InfoMap and signed Leiden RBC at $\gamma \in \{1, 2\}$) yield the same qualitative pattern in the coupling tests below, and we report the full method comparison in Appendix~\ref{app:methods}, Table~\ref{tab:community_method_comparison_overall}.

\begin{figure}[htbp]
\centering
\includegraphics[width=\linewidth]{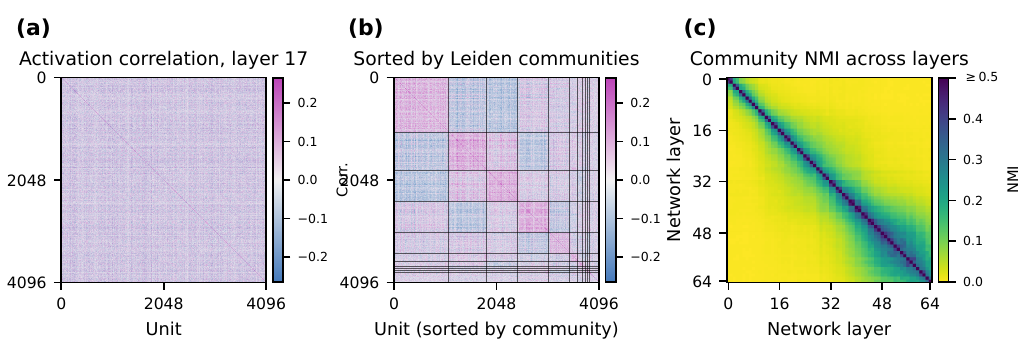}
\caption{Activation-correlation graphs and community structure (OLMo 3 7B, step 1.41M). \textbf{(a)} Pairwise activation correlation matrix at layer~17 (4{,}096 units, original order). \textbf{(b)} Same matrix sorted by signed Leiden CPM communities ($\gamma = 0.001$). \textbf{(c)} Pairwise NMI between community partitions across all sub-layer snapshots (pre-attn even, pre-MLP odd).}
\label{fig:3-1}
\end{figure}

The communities are visually clear (Figure~\ref{fig:3-1}a,b) and reorganize gradually with depth: the pairwise normalized mutual information (NMI) between partitions decays smoothly with layer distance (Figure~\ref{fig:3-1}c), with no discrete phase boundaries. 
% This continuity matters for the analyses to follow---it lets us treat depth as a continuous axis along which both topology (the partition itself) and dynamics (the per-layer Jacobian) evolve, rather than as a sequence of qualitatively distinct phases.

\subsection{Communities capture Jacobian variance even at initialization---a partly architectural prior}
\label{sec:topology-mesoscale}

We reasoned that if topology and dynamics are coupled, the layers where the community structure changes the most may also show the largest dynamical shifts. 
They largely do not---the correlation between adjacent-layer topology disruption ($1 - \text{NMI}$) and changes in the community-projected operator's spectrum is at chance in every model and checkpoint (Appendix~\ref{app:rate-level}).

By contrast, community participation (defined as the coarse-grained operator $K = C_{\text{out}}^\top J\, C_{\text{in}}$, where  $C$ is the community basis) explained significantly more of the Jacobian variance than size-matched random community partitions, with the explained variance rising over the layers (Appendix~\ref{app:methods}, Figure~\ref{fig:mesoscale-bridge-appendix}).

% Instead of asking whether \textit{changes} in topology track \textit{changes} in dynamics, we ask whether the per-layer partition (i.e. community membership) itself is a useful basis for that layer's Jacobian. 
% We project each Jacobian onto the community basis $C$ (each column encodes one community's membership) to obtain a coarse-grained operator $K = C_{\text{out}}^\top J\, C_{\text{in}}$, and compare the fraction of Jacobian variance the projection retains against the same statistic for $100$ size-matched random partitions (Appendix~\ref{app:methods}). 
% The community basis exceeds the size-matched null ($z > 1.96$) at $27/32$ layers in Llama (median $z = 4.41$), $32/32$ at both trained OLMo checkpoints (median $z = 7.74$ and $6.73$), and $21/42$ in Gemma (median $z = 2.31$, with the lower fraction reflecting Gemma's weaker community structure in early/mid layers).

This was true even at OLMo's step 0: the untrained network showed $32/32$ layers above null with median $z = 19.2$---the largest effect of any configuration we tested. 
The residual-stream wiring and attention/MLP block topology are by themselves sufficient to make community structure dynamically informative; mesoscale topology$\leftrightarrow$dynamics agreement is therefore a partly architectural prior, not a learned property. 

\subsection{Training installs a per-unit coupling between community boundaries and Jacobian amplification}
\label{sec:topology-perunit}
Our mesoscale experiments showed that on a layer by layer level, the transformer architecture connects residual stream network topology to its dynamics. This does not, however, address the question of whether topology has any effect on particular residual stream dimensions or units.

\begin{figure}[!t]
\centering
\includegraphics[width=\linewidth]{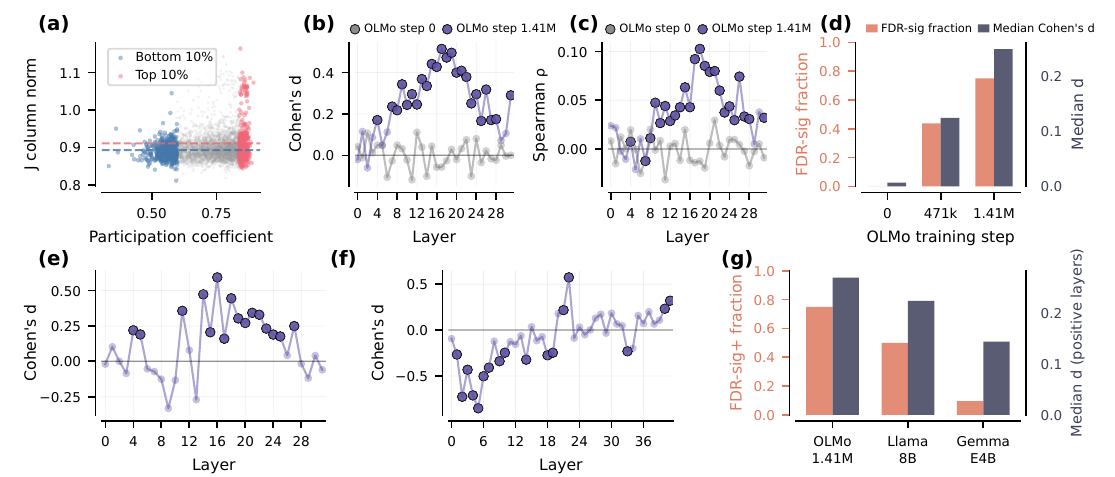}
\caption{Boundary-node coupling: training trajectory and cross-architecture generalization. \textbf{Top row} (OLMo): \textbf{(a)}~participation coefficient vs.\ Jacobian column-norm (layer~17, step~1.41M); \textbf{(b)}~per-layer Cohen's $d$, step~0 (gray) vs.\ step~1.41M (purple); \textbf{(c)}~Spearman $\rho$ by layer; \textbf{(d)}~OLMo training trajectory of FDR-significant fraction and median $d$. \textbf{Bottom row} (cross-model): \textbf{(e)}~Llama 3.1 8B and \textbf{(f)}~Gemma 4 E4B per-layer Cohen's $d$ (darker filled markers are FDR-significant); Gemma shows the sign inversion in non-normal early/mid layers. \textbf{(g)}~Cross-model FDR-significant positive fraction.}
\label{fig:boundary-node}
\end{figure}

Network topology allowed us to measure how evenly a residual stream unit's connections spread across communities using each unit's participation coefficient: high participation means the unit bridges multiple communities (a boundary node), low participation means its connections concentrate within one community. 
We asked whether boundary units are preferentially amplified by the Jacobian, where ``amplification'' is read off the column norm $\|J_{:,i}\|$ (Figure~\ref{fig:boundary-node}a). 
Significance was assessed per layer with Cohen's $d$ between the column norms of units in the top-10\% versus bottom-10\% participation tails, FDR-corrected (Benjamini--Hochberg, $\alpha = 0.05$) across layers within each model; full procedure in Appendix~\ref{app:methods}. Whereas at OLMo step~0, the coupling is null ($0 / 32$ layers reach FDR significance), training installs the coupling monotonically: by step~471k it is detectable at $14 / 32$ layers (median $d = +0.124$), and by step~1.41M at $24 / 32$ layers (median $d = +0.248$) (Figure~\ref{fig:boundary-node}b,d). 
% Whatever wires up the mesoscale prior at initialization---residual-stream architecture, attention/MLP topology---does not extend to per-unit amplification heterogeneity. 
% That heterogeneity is built by training.

% The per-unit effect sizes underlying these counts are small to medium ($d = 0.12$--$0.25$ in Llama and trained OLMo), so the trajectory $0 \to 14 \to 24$ is a statement about \textit{how broadly across depth} the coupling becomes detectable rather than about its magnitude at any single layer. 
% The step-0 baseline median is $+0.007$, however---statistically indistinguishable from zero, not a weak-but-positive value---so what training installs is genuinely new structure rather than amplification of an existing prior.

Llama 3.1 8B reproduces the trained pattern at $16 / 32$ FDR-significant layers (median $d = +0.131$; Figure~\ref{fig:boundary-node}e). 
Gemma 4 E4B replicated the broad picture but with a twist: Only $4 / 42$ layers show FDR-significant positive coupling, while $13 / 42$ show FDR-significant \textit{negative} coupling, the latter clustered in the strongly non-normal early and mid layers where boundary units are preferentially \textit{de}-amplified (Figure~\ref{fig:boundary-node}f).
This sign inversion motivated us to investigate the relationship between boundary-node coupling to dynamics and properties of the Jacobians. 

% \subsection{Operator type governs the sign of boundary node coupling to dynamics}
% \label{sec:topology-mechanism}

In Llama and OLMo, the per-unit coupling is uniformly positive and concentrates in the mid-to-late layers. 
In Gemma, it is positive in the four near-symmetric final layers and negative in the strongly non-normal early and mid layers. 
However, both patterns are instances of the same monotone relationship: the per-layer self-alignment of $J_\ell$---the operator-type measure of \S\ref{sec:gradient}---predicts both the \textit{existence} and the \textit{sign} of the boundary-node coupling.

\begin{figure}[!t]
\centering
\includegraphics[width=\linewidth]{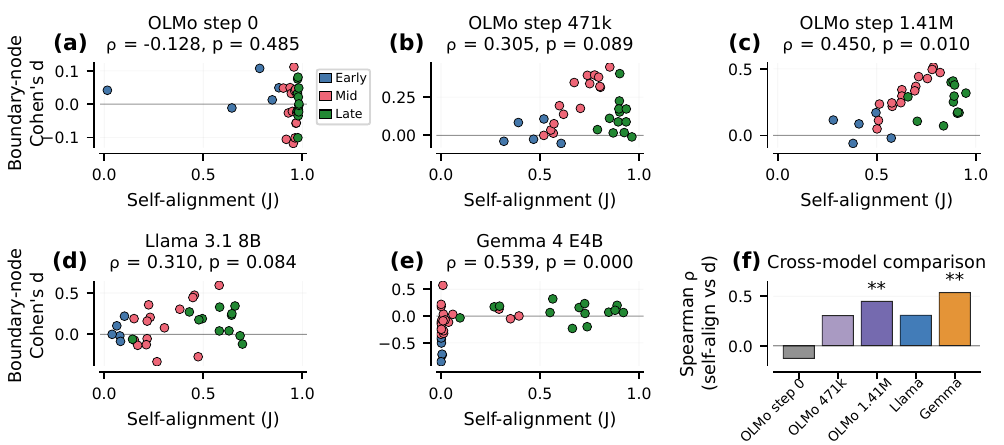}
\caption{Self-alignment (operator type) predicts the sign and magnitude of boundary-node coupling across models and training stages. \textbf{Top row:} OLMo training trajectory --- \textbf{(a)}~step~0, \textbf{(b)}~step~471k, \textbf{(c)}~step~1.41M. \textbf{Bottom row:} \textbf{(d)}~Llama 3.1 8B; \textbf{(e)}~Gemma 4 E4B, whose widest self-alignment range reveals the full monotone relationship including the sign inversion at low self-alignment; \textbf{(f)}~cross-model Spearman $\rho$ summary (${**} = p < 0.01$). Per-panel Spearman $\rho$ and $p$ annotated; points colored by depth regime (blue early, red mid, green late).}
\label{fig:selfalign-vs-boundary}
\end{figure}

Across all five configurations tested, the per-layer Spearman correlation between self-alignment and Cohen's $d$ tracks a single emerging pattern (Figure~\ref{fig:selfalign-vs-boundary}), a monotonic relationship, where layers with self-alignment below $\sim 0.1$ have $d < 0$ (boundary de-amplification), layers above $\sim 0.5$ have $d > 0$ (boundary amplification), and the crossover lies near the operator-type transition between non-normal and near-symmetric. 
 These results demonstrate taht training does not distribute amplification uniformly across residual stream units but concentrates it (or, in non-normal layers, suppresses it) at the units that bridge distinct functional communities.

\section{Discussion}
\label{sec:discussion--limitations--and-future-work}

Dynamical systems theory has proven a productive lens for understanding neural networks, yet it has so far been applied almost entirely at initialization or in toy regimes. 
Network topology offers a complementary lens, with community detection on activation graphs revealing mesoscale functional organization, but whether such organization relates to model dynamics has not been established. 
Here we investigate the full spectral geometry of three production-scale LLMs across training, and connect their dynamics to the residual stream's community topology. 
Three main findings emerge: First, a monotonic non-normality gradient runs from rotation-dominated early layers to near-symmetric late layers, a depth-varying operator not detectable from singular value analyses alone.
Second, composing these operators end-to-end reveals a cumulative low-rank bottleneck that funnels perturbations into a small number of effective channels.
Third, boundary nodes in the activation-graph community structure are preferentially coupled to the Jacobian dynamics, with a sign that is a learned function of operator type: boundary units are amplified at near-symmetric layers de-amplified at non-normal ones. 
Together, these results provide the first full eigenvalue-level description of production-scale transformer Jacobians after training, filling the gap between initialization-time theory, which predicts none of this structure, and approximate linearizations, which cannot resolve the rotational dynamics that dominate the spectrum.

Several methodological choices shape these claims, and each carries a different scope of generalization. The community-detection backbone is signed Leiden CPM at $\gamma = 0.001$, selected on partition-stability grounds; three alternative methods yield the same qualitative pattern across all three topology findings, with CPM giving the largest magnitudes---we therefore read the trained-Jacobian topology--dynamics relationship as algorithm-independent in shape, even where its absolute scale is detector-specific. The per-unit effect sizes themselves are moderate, so the OLMo trajectory of $0 \to 14 \to 24$ FDR-significant layers should be read as a statement about how broadly across depth the coupling becomes detectable rather than about its strength at any single layer; whether prolonged training pushes effect sizes higher or merely propagates the existing signal across more layers is a question the three used OLMo checkpoints cannot settle. Finally, every Jacobian here is evaluated on last-token activations from WikiText-2, so distributional shift to other inputs could in principle affect the results, although we believe the quality of our findings would not be affected. 

Despite these limitations, this work provides useful insights into the dynamics and topology of the transformer residual stream. The dimensional collapse we measure converges with what other groups have found using very different mathematical objects: \citet{golden2025equivalent} report cumulative stable rank of $\sim$1--3 in Llama 3.2 3B via detached Jacobians; \citet{fu2025cast} document a compression--expansion cycle in regression-fit linear maps; \citet{jacobs2025block} observe collapsing stable rank of layer updates in DINOv2-Giant via Dynamic Mode Decomposition. All of these methods and architecture families point at the same progressive funneling through depth, and the exact-Jacobian description we provide here closes the gap that each of those approximate analyses had to leave open. For activation-graph and community-detection methodology, the partition that captures more Jacobian variance than chance is already in place at random initialization: studies that read community structure as a learned organization without an untrained baseline risk attributing to data what is actually due to wiring, and we recommend that activation-graph papers report a random-init baseline whenever the comparison is feasible. For signal-propagation theory, the same step-0 comparison sharpens where the theory still applies: late-regime near-symmetric operators, adjacent-layer SV decoupling at the $\sim$$k/d$ baseline, and the existence of a mesoscale variance signal are all present at initialization. The non-normality gradient, the cumulative dimensional collapse, and the boundary-unit coupling, by contrast, are absent at step~0 and emerge over training---these are exactly the places where initialization-time analysis breaks down and a trained-Jacobian description like the one we provide becomes necessary.

\newpage
\bibliographystyle{plainnat}
\bibliography{references}

%%%%%%%%%%%%%%%%%%%%%%%%%%%%%%%%%%%%%%%%%%%%%%%%%%%%%%%%%%%%
\appendix
\renewcommand{\thefigure}{S\arabic{figure}}
\setcounter{figure}{0}

\section{Methods}
\label{app:methods}
\label{sec:methods}

\subsection{Models and data}

We study three decoder-only transformer families:

\begin{itemize}
  \item \textbf{Llama 3.1 8B} \citep{llama3}: \texttt{meta-llama/Llama-3.1-8B} (final checkpoint). 32 layers, $d = 4{,}096$, 32 attention heads with grouped-query attention (8 KV heads).
  \item \textbf{OLMo 3 7B} \citep{olmo}: \texttt{allenai/Olmo-3-1025-7B} at three training stages (within the pretraining regime): step~0 (random initialization), step~471{,}000, and step~1{,}413{,}814 (final pretraining step). 32 layers, $d = 4{,}096$, 32 attention heads with full multi-head attention.
  \item \textbf{Gemma 4 E4B}: \texttt{google/gemma-4-e4b-it}. 42 layers, $d = 2{,}560$, 10 attention heads with grouped-query attention (2 KV heads). Analyzed at the final (instruction-tuned) checkpoint to test cross-architecture generality.
\end{itemize}

For each configuration, we extract activations on 1{,}000 samples from WikiText-2 \citep{wikitext} (train split, character-length filtered, taken in dataset order). At each transformer block we record the residual-stream state at two sub-layer boundaries---before attention and before the MLP---yielding 64 (Llama, OLMo) or 84 (Gemma) layer snapshots per sample. We retain only the \textbf{last token} position, producing data tensors of shape $(1{,}000 \times 64 \times 4{,}096)$ for Llama/OLMo and $(1{,}000 \times 84 \times 2{,}560)$ for Gemma.

\subsection{Jacobian computation}

The per-layer Jacobian $J_\ell = \partial \mathbf{h}_{\ell+1} / \partial \mathbf{h}_\ell$ is the $d \times d$ linear map ($d = 4{,}096$ for Llama/OLMo, $d = 2{,}560$ for Gemma) describing how an infinitesimal perturbation at the input of transformer block~$\ell$ propagates to its output. Because the block includes a residual connection, $J_\ell = I + \partial f_\ell / \partial \mathbf{h}_\ell$.

We compute exact Jacobians via \texttt{torch.autograd.functional.jacobian}, passing each sample's last-token activation through an isolated copy of the transformer block (cast to float32). Position embeddings are supplied at position~0 for a single-token context; because the computation involves single-token self-attention without a KV cache, the RoPE rotation applied identically to $Q$ and $K$ cancels in the attention dot product, making the result position-invariant.

For each of the 32 layers, we compute 1{,}000 individual Jacobians (one per sample) and define the \textbf{mean Jacobian} $\bar{J}_\ell = \frac{1}{N}\sum_{i=1}^{N} J_\ell^{(i)}$ (accumulated in float64, stored as float32). Per-sample singular value decompositions provide distributional statistics (condition number, Frobenius norm); eigenvalues are computed on the mean Jacobian via \texttt{numpy.linalg.eigvals}.

\subsection{Spectral characterization}

We derive the following quantities from each layer's Jacobians. Together, these capture how each layer stretches, rotates, and compresses the residual stream, and how these transformations compose across depth.

\textbf{Singular values} (per-sample): full SVD of each $J_\ell^{(i)}$. Condition number $\kappa = \sigma_{\max}/\sigma_{\min}$; participation ratio $\text{PR} = (\sum \sigma_i^2)^2 / \sum \sigma_i^4$. The condition number quantifies how anisotropically a layer stretches its input: large $\kappa$ means the layer nearly annihilates some directions while amplifying others, a signature of low-rank or feature-selective computation. The participation ratio complements this by measuring how many singular values carry appreciable weight---a layer that distributes energy across many directions (high PR) operates differently from one that concentrates it in a few (low PR), even if both share the same condition number.

\textbf{Eigenvalues} (mean Jacobian): fraction of expanding modes $= |\{i : |\lambda_i| > 1\}| / d$. Complex conjugate pairs identified at threshold $|\text{Im}(\lambda)| > 10^{-6}$. The fraction of expanding modes indicates whether a layer, on average, amplifies or contracts the residual stream. Complex eigenvalue pairs signal rotational dynamics: the layer mixes pairs of directions rather than simply scaling them. A layer with many complex pairs implements a qualitatively different transformation---one that rotates information between subspaces---compared to a layer whose eigenvalues are predominantly real.

\textbf{Self-alignment}: truncated SVD ($k = 64$) of $\bar{J}_\ell = U \Sigma V^\top$. Self-alignment $= \|V_k^\top U_k\|_F^2 / k$; equals 1 for normal matrices, near $k/d \approx 0.016$ for random subspaces. Normal matrices have orthogonal eigenvectors and their spectral decomposition fully determines their behavior. Non-normal matrices---those with low self-alignment---can transiently amplify inputs even when all eigenvalues indicate contraction, because their input and output directions are misaligned. This transient amplification is invisible to eigenvalue analysis alone and has direct consequences for how perturbations propagate through the network.

\textbf{Forward alignment}: $\|U_\ell^\top V_{\ell+1}\|_F^2 / k$ measures whether adjacent layers compose through their leading directions. When forward alignment is high, the output subspace of layer $\ell$ feeds directly into the input subspace of layer $\ell+1$, meaning information flows efficiently between layers without being scattered into directions the next layer ignores. Low forward alignment suggests that inter-layer composition is indirect, relying on secondary singular directions or the residual stream's skip connection to preserve information.

\textbf{Henrici departure}: $\delta(J) = \sqrt{\|J\|_F^2 - \sum|\lambda_i|^2}\,/\,\|J\|_F$. This is a scalar summary of non-normality derived from the gap between the Frobenius norm and the eigenvalue spectrum. It equals zero for normal matrices and approaches one when the matrix's behavior is dominated by its off-diagonal structure. We use it alongside self-alignment because the two measures are sensitive to different aspects of non-normality: self-alignment captures directional misalignment of the leading singular subspaces, while the Henrici departure reflects the aggregate energy in the strictly upper-triangular part of the Schur decomposition.

\textbf{Residual operator} $R_\ell = J_\ell - I$: the same alignment and spectral analyses applied after stripping the skip connection. In a residual architecture, $J_\ell = I + R_\ell$, so the identity contributes a baseline of unit singular values and real eigenvalues at one. Analyzing $R_\ell$ isolates what the attention and MLP sublayers actually compute---the perturbation to the residual stream---from the pass-through signal. This decomposition is necessary because the skip connection can mask the spectral structure of the learned transformation: a layer whose $J_\ell$ appears well-conditioned may have an $R_\ell$ that is highly anisotropic.

\textbf{Cumulative Jacobian} $P_\ell = J_{31} \cdots J_\ell$: computed via iterative backward SVD composition (truncated at $k = 512$). Effective rank $= \exp(-\sum p_i \log p_i)$ where $p_i = \sigma_i^2 / \sum \sigma_j^2$. The cumulative Jacobian captures the end-to-end linear sensitivity of the output to perturbations at layer $\ell$, integrating the effects of all downstream layers. Its effective rank measures the dimensionality of the subspace through which information from layer $\ell$ can influence the final representation. A sharp drop in effective rank with depth indicates a computational bottleneck where the network progressively discards degrees of freedom.

\subsection{Activation-correlation graphs and community detection}

For each of the 64 sub-layer snapshots, we construct a sparse signed correlation graph:
\begin{enumerate}
  \item Z-score each unit across the 1{,}000 samples.
  \item Compute the full $4{,}096 \times 4{,}096$ Pearson correlation matrix.
  \item Retain the top-$k = 20$ edges per node (by $|\text{corr}|$), symmetrize by keeping the larger absolute weight, exclude self-loops.
\end{enumerate}
Edge signs (positive and negative correlations) are retained. We detect communities using \textbf{signed Leiden CPM} \citep{traag2019louvain} with resolution $\gamma = 0.001$ on the positive subgraph and $\gamma_{\text{neg}} = 0$ on the negative subgraph, optimized jointly with layer weights $[1, -1]$. This yields $6$--$87$ non-degenerate communities per layer across all configurations.

\begin{table}[t]
  \centering
  \small
  \caption{Comparison of community-detection methods on sparse signed activation-correlation graphs (top-$k=20$ edges per node by $|\mathrm{corr}|$), aggregated over 64 sub-layer snapshots per model. \emph{Frac.\ FDR sig.\ ($+$)} is the fraction of layers with a false-discovery-rate significant positive bridge effect; Cohen's $d$ is summarized by the median and mean across layers.}
  \label{tab:community_method_comparison_overall}
  \setlength{\tabcolsep}{5pt}
  \begin{tabular}{@{}llccc@{}}
    \toprule
    Model & Method & \begin{tabular}[c]{@{}c@{}}Frac.\ FDR\\sig.\ ($+$)\end{tabular} & \begin{tabular}[c]{@{}c@{}}Median\\Cohen's $d$\end{tabular} & \begin{tabular}[c]{@{}c@{}}Mean\\Cohen's $d$\end{tabular} \\
    \midrule
    \multirow{4}{*}{\shortstack{OLMo3-7B\\step 0 (untrained)}}
      & $|\mathrm{corr}|$ InfoMap & 0.000 & 0.022 & 0.023 \\
      & signed Leiden $\gamma{=}1$ & 0.000 & 0.013 & 0.009 \\
      & signed Leiden $\gamma{=}2$ & 0.000 & $-0.008$ & $-0.006$ \\
      & signed Leiden CPM $\gamma{=}0.001$ & 0.000 & 0.007 & 0.003 \\
    \addlinespace
    \multirow{4}{*}{\shortstack{OLMo3-7B\\step 471k}}
      & $|\mathrm{corr}|$ InfoMap & 0.094 & 0.046 & 0.048 \\
      & signed Leiden $\gamma{=}1$ & 0.125 & 0.102 & 0.094 \\
      & signed Leiden $\gamma{=}2$ & 0.281 & 0.119 & 0.112 \\
      & signed Leiden CPM $\gamma{=}0.001$ & 0.438 & 0.124 & 0.160 \\
    \addlinespace
    \multirow{4}{*}{\shortstack{OLMo3-7B\\step 1.4M\\(trained)}}
      & $|\mathrm{corr}|$ InfoMap & 0.125 & 0.068 & 0.075 \\
      & signed Leiden $\gamma{=}1$ & 0.250 & 0.119 & 0.120 \\
      & signed Leiden $\gamma{=}2$ & 0.375 & 0.135 & 0.147 \\
      & signed Leiden CPM $\gamma{=}0.001$ & 0.750 & 0.248 & 0.254 \\
    \addlinespace
    \multirow{4}{*}{Llama-3.1-8B}
      & $|\mathrm{corr}|$ InfoMap & 0.469 & 0.110 & 0.149 \\
      & signed Leiden $\gamma{=}1$ & 0.031 & 0.027 & 0.023 \\
      & signed Leiden $\gamma{=}2$ & 0.094 & 0.054 & 0.038 \\
      & signed Leiden CPM $\gamma{=}0.001$ & 0.500 & 0.131 & 0.116 \\
    \bottomrule
  \end{tabular}
\end{table}

The \textbf{participation coefficient} of unit $i$ is $p_i = 1 - \sum_c (k_{ic} / k_i)^2$, where $k_{ic}$ is the sum of absolute edge weights from $i$ to community $c$ and $k_i = \sum_c k_{ic}$.

\subsection{Statistical tests}

\paragraph{Test 1 (rate-level coupling).} Spearman correlation between topology disruption ($1 - \text{NMI}$ of adjacent-layer communities) and $\Delta$~dynamics ($\Delta \sigma_{\max}(K)$ or $|\Delta\text{variance captured}|$); $n = 31$ layer pairs. Significance via 10{,}000-permutation two-sided test.

\paragraph{Test 2 (mesoscale variance captured).} The mesoscale operator $K = C_{\text{out}}^\top J C_{\text{in}}$ projects the Jacobian onto community bases ($C$: columns are $1/\sqrt{n_c}$ for community members, zero elsewhere). Variance captured $= \|C_{\text{out}} K C_{\text{in}}^\top\|_F^2 / \|J\|_F^2$, compared against 100 random same-size partitions (one-sided $z$-test).

\paragraph{Test 3 (boundary-node amplification).} Per layer: Cohen's $d$ between Jacobian column-norm $\|J_{:,i}\|_2$ for units in the top-10\% vs.\ bottom-10\% participation tails (denominator: $\text{std}(\text{all column norms}, \text{ddof}=1)$). Significance via 5{,}000-permutation two-sided test; FDR correction (Benjamini--Hochberg) across 32 layers within each model at $\alpha = 0.05$.

\subsection{Software}

All computations use Python~3.12, PyTorch~$\geq$2.5 (CUDA~12.4), HuggingFace Transformers~$\geq$5.2, nnsight~$\geq$0.4 for activation extraction, and leidenalg for community detection. Code is available at \texttt{[redacted for review]}.

\begin{figure}[!t]
\centering
\includegraphics[width=\linewidth]{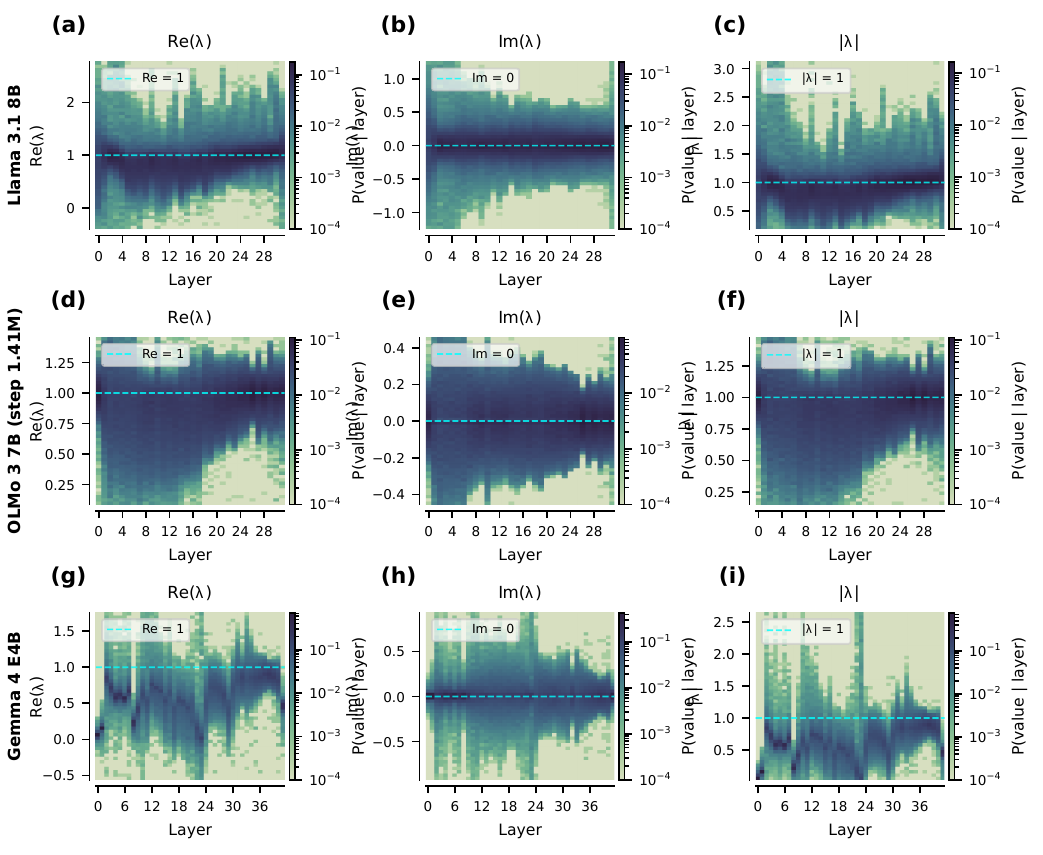}
\caption{Eigenvalue distribution heatmaps for all three trained models, organized as a 3$\times$3 grid. Rows are models (\textbf{(a--c)}~Llama 3.1 8B; \textbf{(d--f)}~OLMo 3 7B step~1.41M; \textbf{(g--i)}~Gemma 4 E4B); columns are Re($\lambda$), Im($\lambda$), and $|\lambda|$. Color encodes $P(\text{value} \mid \text{layer})$ (log scale). Dashed lines mark Re $= 1$, Im $= 0$, and $|\lambda| = 1$. All three models show the same three-regime structure: broad early distributions, contracted mid-layers, and re-expanding late layers.}
\label{fig:eigenvalue-heatmaps-appendix}
\end{figure}

\begin{figure}[!t]
\centering
\includegraphics[width=\linewidth]{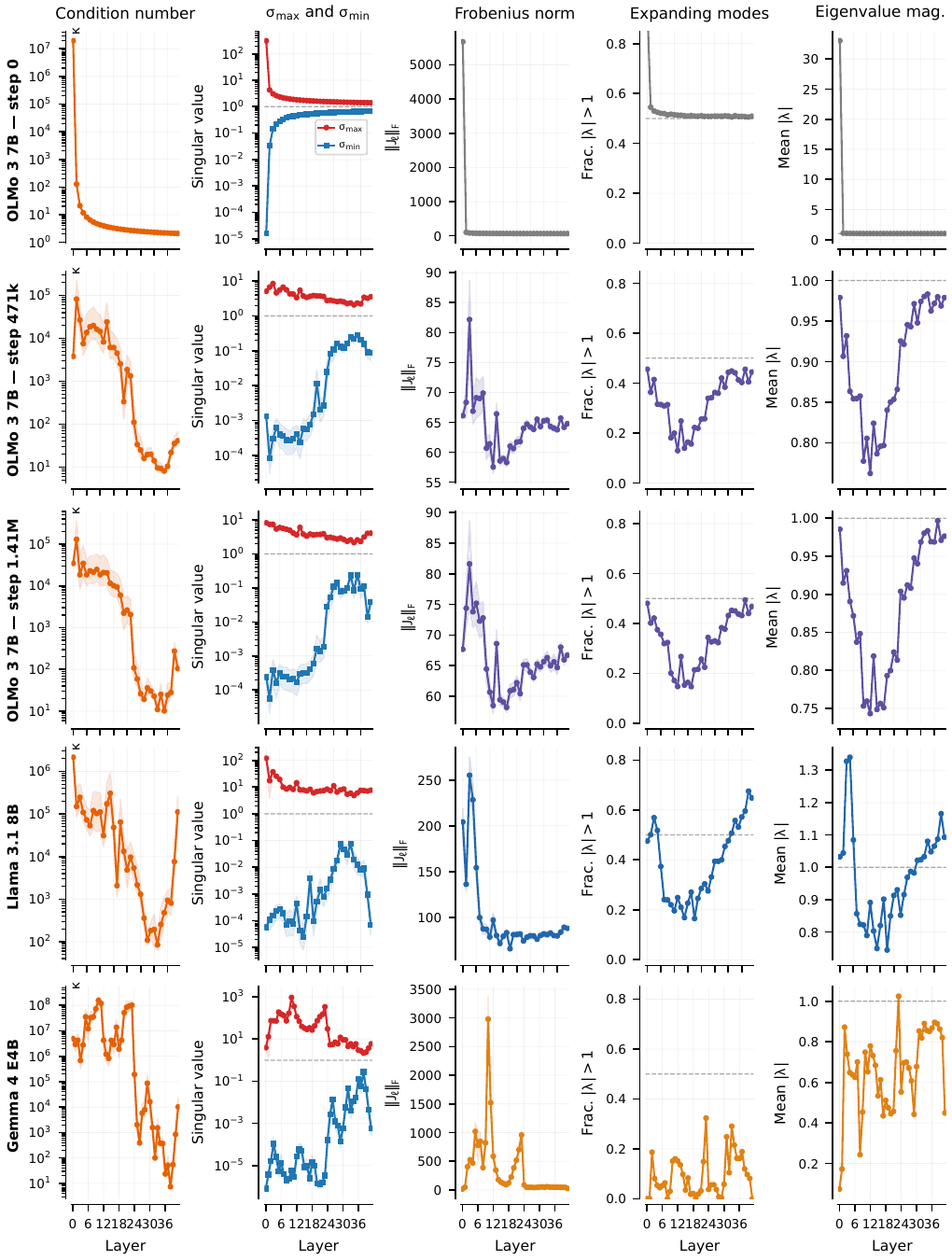}
\caption{Per-layer spectral profiles for all five configurations (companion to Figure~\ref{fig:section2-1}). Each row is one model/checkpoint (top to bottom: OLMo step~0, OLMo step~471k, OLMo step~1.41M, Llama 3.1 8B, Gemma 4 E4B); columns show, from left to right: condition number $\kappa = \sigma_{\max}/\sigma_{\min}$, the leading and trailing singular values $\sigma_{\max}$ and $\sigma_{\min}$, the Frobenius norm $\|J_\ell\|_F$, the fraction of expanding eigenvalue modes ($|\lambda|>1$), and the mean eigenvalue magnitude. $\sigma_{\max}$ is comparatively stable across mid/late layers in every trained model, while $\sigma_{\min}$ collapses at both extremes---the singular-value mechanism behind the U-shaped condition number profile.}
\label{fig:section2-1-appendix}
\end{figure}

\begin{figure}[!t]
\centering
\includegraphics[width=\linewidth]{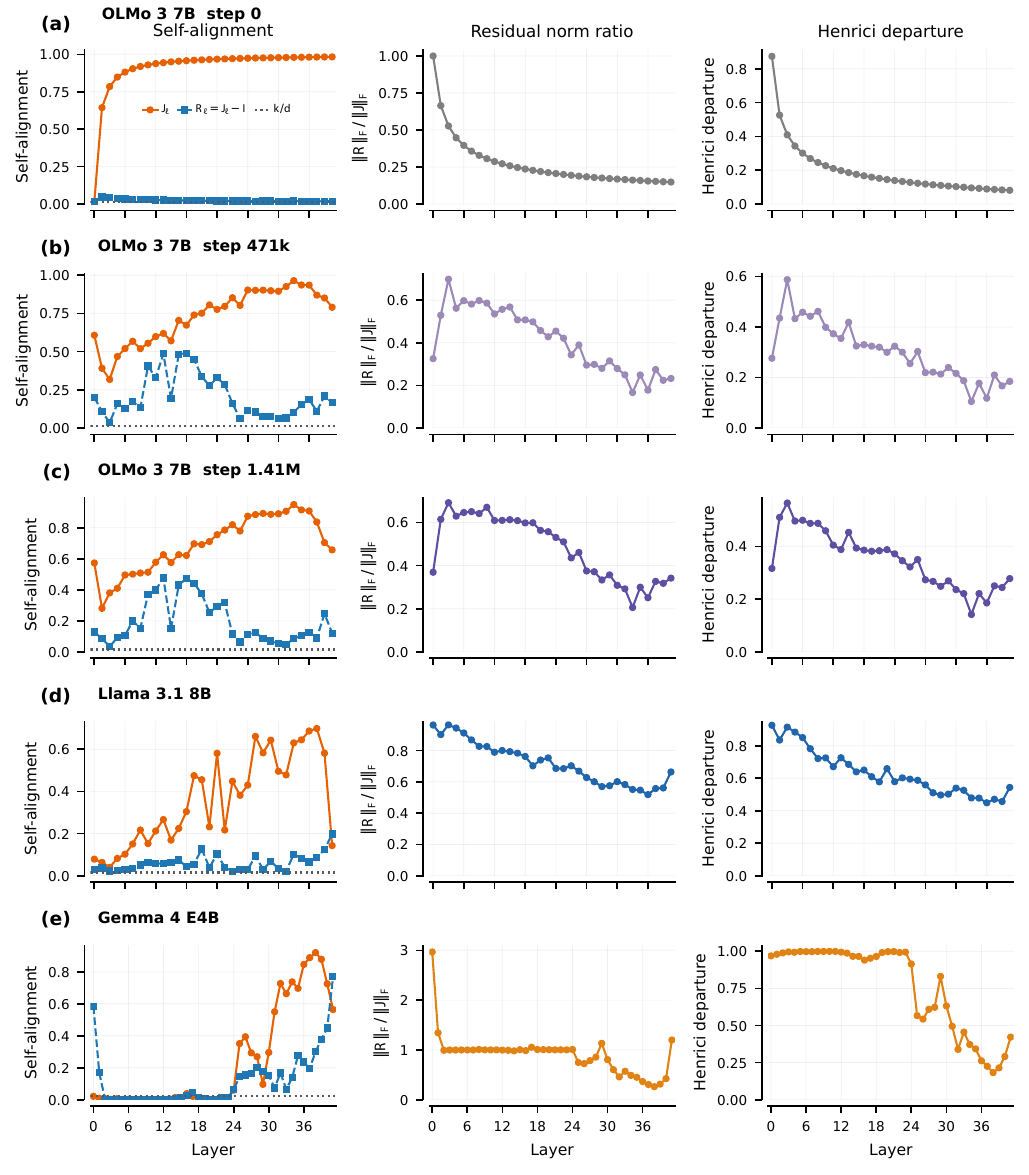}
\caption{Per-layer non-normality metrics for all five configurations (companion to Figure~\ref{fig:non-normality-composite}). Each row is one model/checkpoint. \textit{Left:} Self-alignment of $J$ (orange) and $R$ (blue) with $k/d$ baseline. \textit{Center:} Residual norm ratio. \textit{Right:} Henrici departure. The non-normality gradient (low early, high late) is present in all trained models; Gemma spans the widest range (0.0003--0.92). OLMo step~0 saturates to near-symmetric by layer~5.}
\label{fig:nonnormality-appendix}
\end{figure}

\begin{figure}[!t]
\centering
\includegraphics[width=\linewidth]{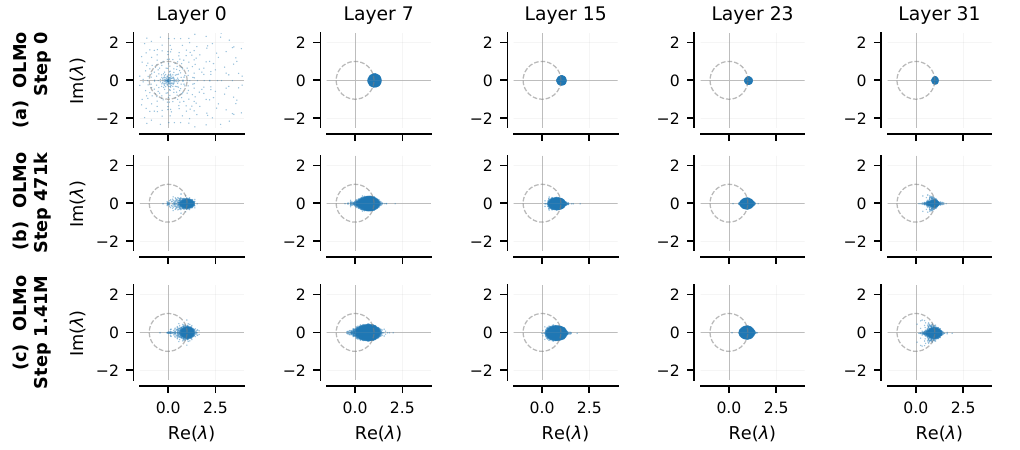}
\caption{Architecture vs.\ training in eigenvalue geometry (OLMo 3 7B). Each row shows eigenvalues at layers 0, 7, 15, 23, 31 (dashed circle = unit circle). \textbf{(a)}~Step~0: eigenvalues collapse to a tight cluster near $\mathrm{Re} = 1$ at all layers beyond layer~0. \textbf{(b)}~Step~471k: structured elliptical clouds emerge, with expanding mid-layer modes. \textbf{(c)}~Step~1.41M: rich spectral diversity at every layer, with persistent complex-plane geometry matching Llama's trained structure. Full eigenvalue distribution heatmaps for all models are in Figure~\ref{fig:eigenvalue-heatmaps-appendix}.}
\label{fig:2-1e}
\end{figure}

\begin{figure}[!t]
\centering
\includegraphics[width=\linewidth]{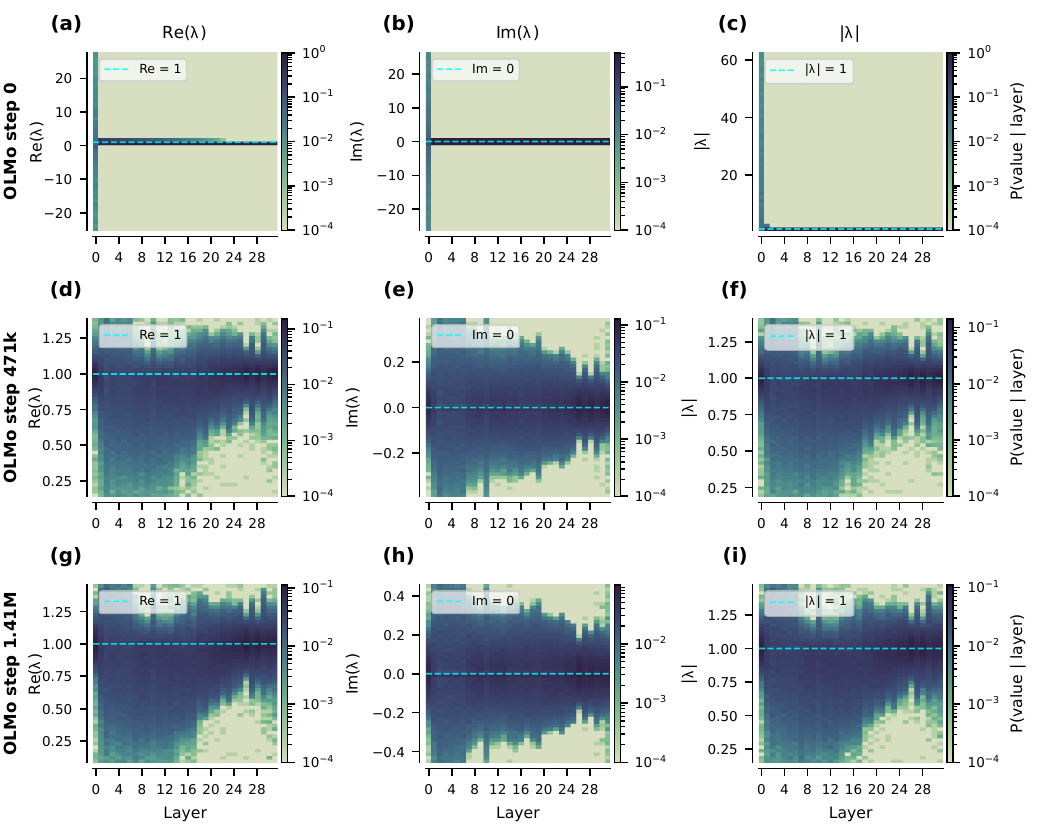}
\caption{Eigenvalue distribution heatmaps across OLMo training (companion to Figure~\ref{fig:2-1e}). \textbf{(a--c)}~Step~0: eigenvalues are tightly concentrated around Re $\approx 1$, Im $\approx 0$, $|\lambda| \approx 1$ with only layer~0 showing spread. \textbf{(d--f)}~Step~471k: structured distributions emerge. \textbf{(g--i)}~Step~1.41M: full three-regime structure matching Llama. Training progressively installs spectral diversity.}
\label{fig:olmo-eigenvalue-heatmaps-appendix}
\end{figure}

\begin{figure}[!t]
\centering
\includegraphics[width=\linewidth]{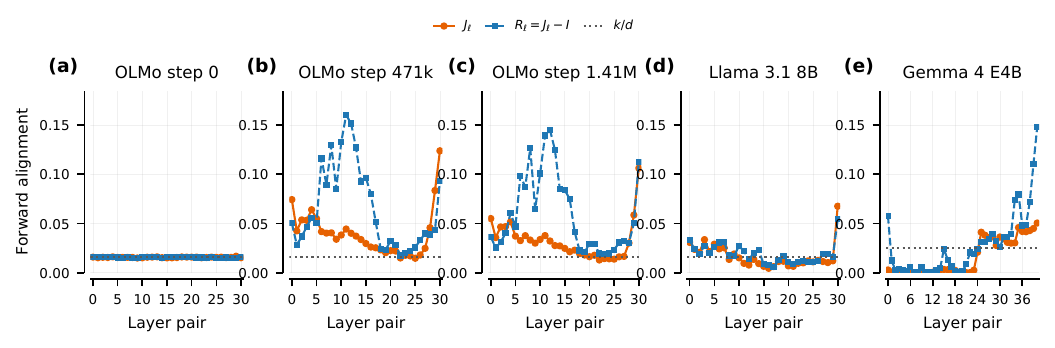}
\caption{Adjacent-layer forward alignment $\|U_\ell^\top V_{\ell+1}\|_F^2 / k$ for $J_\ell$ (orange) and $R_\ell = J_\ell - I$ (blue), with random baseline $k/d \approx 0.016$ (dotted). Panels show one model/checkpoint each: \textbf{(a)}~OLMo step~0, \textbf{(b)}~OLMo step~471k, \textbf{(c)}~OLMo step~1.41M, \textbf{(d)}~Llama 3.1 8B, \textbf{(e)}~Gemma 4 E4B. At step~0, $J$ and $R$ are exactly at baseline---no inter-layer coordination at initialization. By step~471k, $R$ begins to show elevated mid-layer values; at step~1.41M, $R$ shows a pronounced mid-layer peak ($3.8\times$ baseline, layers 6--14), indicating training installs block-level inter-layer coordination in the depth range where the operator-type transition is steepest. Llama shows no such peak; Gemma shows a modest late-layer elevation.}
\label{fig:forward-alignment-appendix}
\end{figure}

\begin{figure}[!t]
\centering
\includegraphics[width=\linewidth]{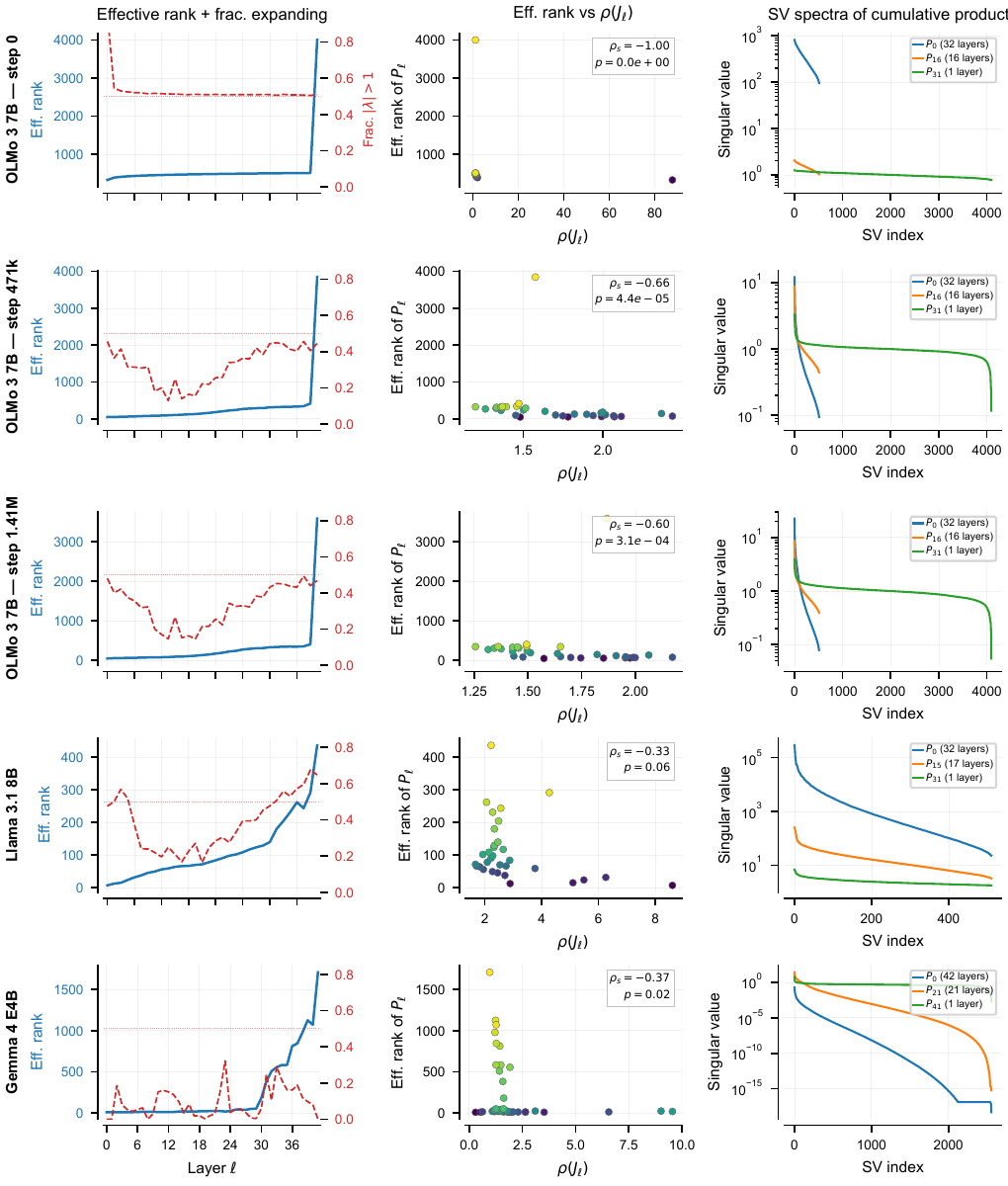}
\caption{Cumulative Jacobian analysis for all five configurations (companion to Figure~\ref{fig:cumulative-jacobian}). Each row is one model/checkpoint (top to bottom: OLMo step~0, OLMo step~471k, OLMo step~1.41M, Llama 3.1 8B, Gemma 4 E4B). \textit{Left:} Effective rank of $P_\ell$ (blue, left axis) and fraction of expanding eigenvalues (red dashed, right axis) vs.\ injection layer. \textit{Center:} Effective rank vs.\ spectral radius $\rho(J_\ell)$, colored by layer; Spearman $\rho_s$ and $p$ annotated. At step~0, spectral radii cluster near 1 and effective rank is uniformly high; training progressively installs the negative correlation. \textit{Right:} Singular value spectra of $P_0$ (full-depth product), $P_{\mathrm{mid}}$, and $P_{L-1}$ (single layer). The trained models show orders-of-magnitude dynamic range concentrated in the leading modes; the untrained model has a nearly flat spectrum. Spectra are shown for the singular values computed in the saved decomposition: Gemma spectra and the single-layer OLMo $P_{31}$ spectra use full SVD, whereas the 4096-dimensional OLMo cumulative products and Llama spectra use the iterative truncated-SVD computation with $k=512$ for tractability; zero padding of truncated spectra is omitted from the plot.}
\label{fig:cumulative-jacobian-appendix}
\end{figure}

\begin{figure}[!t]
\centering
\includegraphics[width=\linewidth]{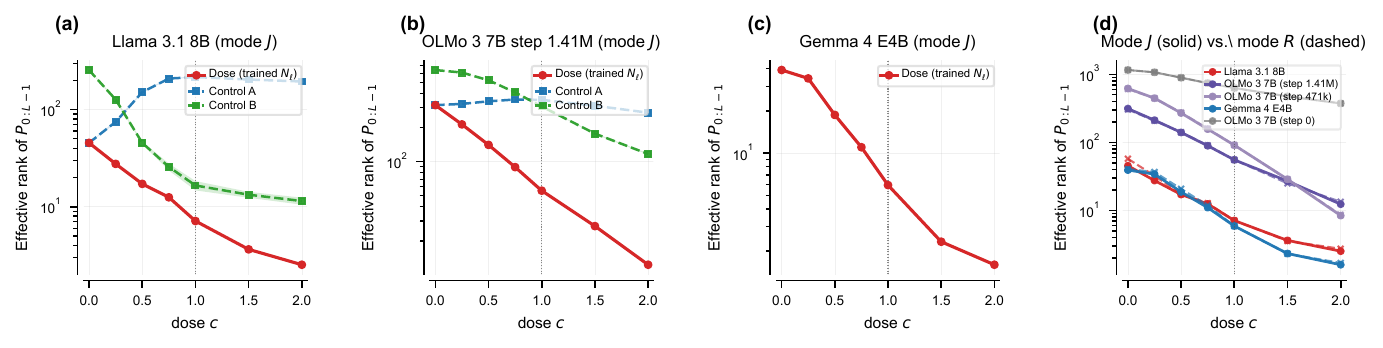}
\caption{Schur surgery: random controls and mode-$R$ confirmation. Cumulative effective rank of $P_{0:L-1}$ vs.\ dose $c$ in mode $J$ for \textbf{(a)}~Llama 3.1 8B and \textbf{(b)}~OLMo 3 7B (step~1.41M): trained dose (red, solid) vs.\ Control~A (Frobenius-matched random replacement of $N_\ell$, blue dashed) and Control~B (Haar-random Schur basis, green dashed); shaded bands show $\pm$std across 4 independent random draws per layer. Control~A reverses the dose response in both networks; Control~B partly recovers the funnel in Llama but not in OLMo. \textbf{(c)}~Trained-dose curve for Gemma 4 E4B (mode $J$). \textbf{(d)}~Mode-$J$ (solid) vs.\ mode-$R$ (dashed) dose curves overlaid for all five configurations: the two modes are nearly indistinguishable, ruling out the $+I$ skip as a source of the funnel.}
\label{fig:schur-surgery-appendix}
\end{figure}

\begin{figure}[!t]
\centering
\includegraphics[width=\linewidth]{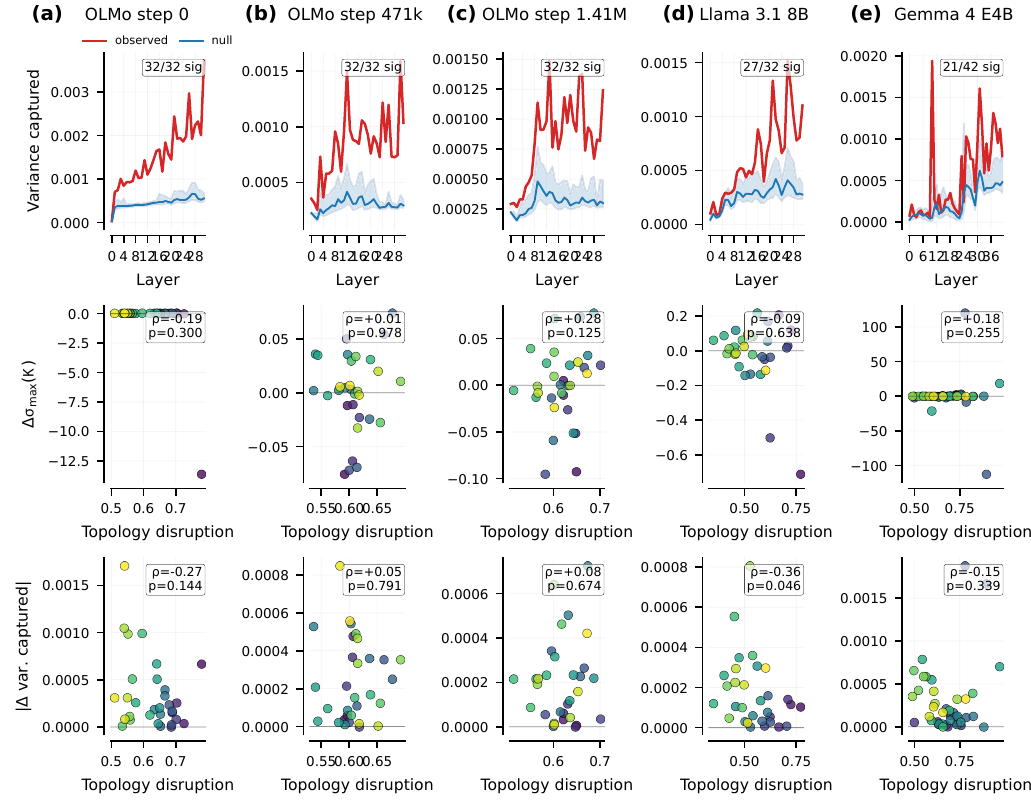}
\caption{Mesoscale bridge diagnostic across all models and training checkpoints (signed Leiden CPM $\gamma = 0.001$). Columns: \textbf{(a)}~OLMo step~0, \textbf{(b)}~OLMo step~471k, \textbf{(c)}~OLMo step~1.41M, \textbf{(d)}~Llama 3.1 8B, \textbf{(e)}~Gemma 4 E4B. \textit{Top row:} Variance captured by the community-projected operator $K$ (red) vs.\ null distribution from 100 random same-size partitions (blue band = 5--95\%); annotation shows the count of layers with $z > 1.96$. \textit{Middle row:} Rate-level coupling --- topology disruption ($1 - \text{NMI}$) vs.\ \emph{signed} $\Delta\sigma_{\max}(K)$; dot color encodes layer depth. The figure annotations show Spearman correlations on the signed measure (e.g., Gemma $\rho = +0.18$, $p = 0.255$); the absolute-value version $|\Delta\sigma_{\max}(K)|$ used in the main text is reported in Table~\ref{tab:rate-level} (Gemma: $\rho = +0.39$, $p = 0.012$). \textit{Bottom row:} Topology disruption vs.\ $|\Delta\text{variance captured}|$; Spearman $\rho$ and $p$-values annotated per panel.}
\label{fig:mesoscale-bridge-appendix}
\end{figure}

\begin{table}[!t]
\centering
\small
\begin{tabular}{lllll}
\toprule
Run & early & mid & late & min layer \\
\midrule
OLMo step 0 (untrained) & 0.635 & 0.951 & 0.979 & layer 0 (0.016) \\
OLMo step 1.41M (final) & 0.428 & 0.653 & 0.859 & layer 1 (0.281) \\
Llama 3.1 8B (final)    & 0.073 & 0.299 & 0.556 & layer 2 (0.040) \\
Gemma 4 E4B (final)     & 0.005 & 0.061 & 0.660 & layer 9 (0.0003) \\
\bottomrule
\end{tabular}
\caption{Mean $J_\ell$ self-alignment by regime (early/mid/late split at layers 5, 20 for 32-layer models; 8, 28 for Gemma's 42 layers). Training broadens the rotator regime in early and mid layers while leaving the late-layer near-symmetry largely intact.}
\label{tab:selfalign-regimes}
\end{table}

\section{Adjacent-layer forward alignment}
\label{app:forward-alignment}

The forward alignment $\|U_\ell^\top V_{\ell+1}\|_F^2 / k$ measures whether the leading output subspace of layer~$\ell$ overlaps with the leading input subspace of layer~$\ell+1$, i.e.\ whether adjacent blocks compose through their dominant singular directions ($k = 64$). A value near the random baseline $k/d \approx 0.016$ means the two subspaces are no more aligned than chance.

For the full Jacobian $J$, forward alignment sits near baseline across all models and training stages (Figure~\ref{fig:forward-alignment-appendix}a--c). Llama 3.1 8B has mean $0.016$ ($1.05\times$ baseline) with a single spike at the final layer ($0.068$) reflecting the readout boundary; OLMo step~0 sits at $0.016$ ($1.00\times$ baseline)---exactly random at initialization---and OLMo step 1.41M reaches $0.032$ ($2.0\times$), with a final-layer spike ($0.106$). Gemma's mean is $0.016$ ($0.64\times$ its $k/d = 0.025$ baseline), again with a final-layer spike ($0.050$, $2.0\times$). Adjacent-layer subspace coupling at the level of the full Jacobian is therefore at or below random across the board.

Stripping the identity via the residual operator $R = J - I$ reveals a more subtle picture. Llama's $R$ forward alignment is $0.019$ ($1.2\times$ baseline)---nearly unchanged from $J$, consistent with the block computations being uncoupled across adjacent layers. OLMo step~0 $R$ sits exactly at baseline ($1.0\times$). But OLMo step 1.41M $R$ rises to $0.060$ ($3.8\times$ baseline), with a pronounced mid-layer peak (layers 6--14 reach $0.09$--$0.15$; Figure~\ref{fig:forward-alignment-appendix}c). Training installs inter-layer coordination in the block computations concentrated in mid-depth---the same regime where the non-normality gradient is steepest. Even so, the absolute values are small (max $0.15$ on a $[0, 1]$ scale): the residual stream still functions primarily as the shared workspace of \citet{elhage2021framework}, not a sequential pipeline. Whether this mid-layer $R$ peak is a general feature of trained transformers or specific to OLMo's architecture and training recipe remains open---Llama's $R$ forward alignment shows no such peak.

\section{Schur surgery: setup, random controls, and mode-$R$ confirmation}
\label{app:schur-surgery}

This appendix expands on the dose-response intervention reported in \S\ref{sec:schur_surgery}: the formal setup, two random-baseline controls that separately ablate the structural and basis components of the trained $N_\ell$, and a mode-$R$ variant that targets the residual operator $R_\ell = J_\ell - I$.

\subsection{Schur form, dose response, and random controls}

For each per-layer mean Jacobian we compute the complex Schur factorization
\[
J_\ell \;=\; Q_\ell \, (\Lambda_\ell + N_\ell) \, Q_\ell^*,
\]
with $Q_\ell$ unitary, $\Lambda_\ell$ diagonal (the eigenvalues), and $N_\ell$ strictly upper-triangular. Because $Q_\ell$ is unitary,
\[
\|N_\ell\|_F^2 \;=\; \|J_\ell\|_F^2 - \sum_i |\lambda_i|^2,
\]
so $\|N_\ell\|_F$ is identical to the Henrici-departure quantity used in \S\ref{sec:gradient} (it is the same non-normality, written in a form we can manipulate). 
We sweep a dose grid $c \in \{0, 0.25, 0.5, 0.75, 1, 1.5, 2\}$ over three constructions, each replacing $J_\ell$ at every layer of the linearized stack:
\begin{itemize}
    \item \textbf{Dose} (the main intervention): $J_\ell(c) = Q_\ell (\Lambda_\ell + c\, N_\ell) Q_\ell^*$. Spectrum and basis preserved; non-normal feedforward scaled. $c=0$ is fully normal at the trained spectrum, $c=1$ is the trained model.
    \item \textbf{Control A (Frobenius-matched random replacement):} $J_\ell^A(c) = Q_\ell \Lambda_\ell Q_\ell^* + M_\ell^{\mathrm{rand}}$, with $M_\ell^{\mathrm{rand}}$ drawn i.i.d.\ Gaussian and rescaled so $\|M_\ell^{\mathrm{rand}}\|_F = c\,\|N_\ell\|_F$. Preserves the spectrum, the trained basis, and the Frobenius mass of the perturbation; destroys the upper-triangular structure of $N_\ell$.
    \item \textbf{Control B (Haar-random Schur basis):} $J_\ell^B(c) = Q^{\mathrm{rand}} (\Lambda_\ell + c\, N_\ell) (Q^{\mathrm{rand}})^*$, with $Q^{\mathrm{rand}}$ a Haar-random unitary. Preserves the spectrum and the upper-triangular shape of $N_\ell$; replaces the trained Schur basis $Q_\ell$.
\end{itemize}
Both controls draw four independent random matrices per layer ($n=4$). 
The cumulative product $P_{0:\ell}(c) = J_\ell(c) \cdots J_0(c)$ is computed by iterative truncated SVD at rank $K_{\mathrm{cum}} = 512$ with per-layer max-normalization, with a $\log_{10}$ Frobenius scale accumulated separately for the transient-amplification panel of Figure~\ref{fig:schur-surgery}.

\paragraph{Mode $R$.} To confirm the result is not an artifact of the residual identity, we repeat all three constructions on $R_\ell = J_\ell - I$, write $R_\ell = Q_\ell (\Lambda_\ell^R + N_\ell^R) Q_\ell^*$ in Schur form, scale only $N_\ell^R$ by $c$, and substitute $J_\ell(c) = I + R_\ell(c)$ back into the stack.

\subsection{Per-model results}

\begin{table}[!ht]
\centering
\small
\begin{tabular}{lcccc}
\toprule
& \multicolumn{3}{c}{Mode $J$ dose ($\mathrm{erank}\,P_{0:L-1}$)} & Mode $R$ \\
\cmidrule(lr){2-4}
Run & $c=0$ & $c=1$ (trained) & $c=2$ & $c=0/c=1$ \\
\midrule
Llama 3.1 8B            & $45.4$ & $7.12$  & $2.53$ & $7.7\times$ \\
OLMo 3 7B (step 1.41M)  & $314$  & $55.5$  & $12.4$ & $5.6\times$ \\
Gemma 4 E4B             & $39.3$ & $5.92$  & $1.60$ & $6.7\times$ \\
\bottomrule
\end{tabular}
\caption{Schur dose response, cumulative effective rank of $P_{0:L-1}$. Mode $J$ $c=0/c=1$ ratios are $6.4$, $5.7$, $6.6$ across the three trained models; mode $R$ ratios on the residual operator are nearly identical, ruling out the $+I$ skip as a source of the funnel.}
\label{tab:schur-dose}
\end{table}

\paragraph{Random controls at $c=1$ (mode $J$).} Cumulative effective rank of $P_{0:31}$:
\begin{itemize}
    \item \textbf{Llama 3.1 8B:} dose $7.12$; Control~A $215.9 \pm 1.2$ ($n=4$); Control~B $16.5 \pm 1.6$ ($n=4$).
    \item \textbf{OLMo 3 7B (step 1.41M):} dose $55.5$; Control~A $349.4 \pm 0.3$ ($n=4$); Control~B $305.3 \pm 0.8$ ($n=4$).
\end{itemize}
Control~A reverses the dose response in both networks: random feedforward of identical Frobenius mass to the trained $N_\ell$ \emph{raises} cumulative effective rank rather than lowering it (Figure~\ref{fig:schur-surgery-appendix}a). Control~B partly recovers the funnel in Llama (16.5, close to dose 7.1) but not in OLMo (305, close to Control~A): the trained Schur basis $Q_\ell$ is therefore necessary to drive the funnel in OLMo, while the upper-triangular structure of $N_\ell$ alone is largely sufficient in Llama. The bottleneck thus depends on the trained non-normal structure---the upper-triangular shape of $N_\ell$ and, in some architectures, also the trained basis $Q_\ell$ in which it is realized---and not on either the spectrum alone or the Frobenius mass of the off-diagonal perturbation.

\paragraph{Mode $R$ confirmation.} Repeating the surgery on $R_\ell = J_\ell - I$ gives nearly indistinguishable dose curves to mode $J$ at every $c$ across all four configurations (Figure~\ref{fig:schur-surgery-appendix}b; Table~\ref{tab:schur-dose}, last column). The result is therefore a property of the learned block computation, not of the residual identity.

\section{Rate-level topology--dynamics coupling}
\label{app:rate-level}

If topology and dynamics are coupled at the level of layer-to-layer change, then layers where the community structure reorganizes most between consecutive depths should also show the largest swings in dynamical observables. We tested this via Spearman correlation between adjacent-layer topology disruption ($1 - \text{NMI}$ of signed Leiden CPM partitions) and two dynamical measures derived from the community-projected operator $K = C^\top J C$: the change in its dominant singular value $|\Delta \sigma_{\max}(K)|$ and the change in the variance it captures relative to $J$, $|\Delta(\|K\|_F^2 / \|J\|_F^2)|$ ($n = 31$ layer pairs for the 32-layer models, $n = 41$ for Gemma). Table~\ref{tab:rate-level} gives the full per-model statistics; per-model coupling scatters with point colour by layer depth are in the middle and bottom rows of Figure~\ref{fig:mesoscale-bridge-appendix}.

Llama's $|\Delta \text{variance captured}|$ correlation is significantly negative ($\rho = -0.36$, $p = 0.046$): more topology disruption between adjacent layers predicts \textit{smaller} swings in variance captured, the opposite direction to a coupling hypothesis. Gemma's $|\Delta \sigma_{\max}(K)|$ is the only positive coupling, modest in magnitude ($\rho = +0.39$, $p = 0.012$), suggesting that in this deeper network layers where the community partition reshuffles also show larger shifts in mesoscale operator gain. The complementary variance-captured measure is null in Gemma, however, and both measures are null at OLMo step~0 and across the OLMo training trajectory. The aggregate verdict is that rate-level topology--dynamics coupling is not a robust cross-architecture phenomenon, and we do not pursue it further in the main text.

\begin{table}[!t]
\caption{Rate-level coupling tests (signed Leiden CPM $\gamma = 0.001$). $|\Delta\sigma_{\max}(K)|$ and $|\Delta$~variance captured$|$ between adjacent layers correlated with topology disruption ($1 - \text{NMI}$).}
\centering
\small
\begin{tabular}{lll}
\toprule
Run & $|\Delta \sigma_{\max}(K)|$ & $|\Delta$ variance captured$|$ \\
\midrule
Llama 3.1 8B            & $\rho = -0.09$, $p = 0.638$, $n = 31$ & $\rho = -0.36$, $p = 0.046$, $n = 31$ \\
OLMo step 0 (untrained) & $\rho = -0.19$, $p = 0.300$, $n = 31$ & $\rho = -0.27$, $p = 0.144$, $n = 31$ \\
OLMo step 471k          & $\rho = +0.01$, $p = 0.978$, $n = 31$ & $\rho = +0.05$, $p = 0.791$, $n = 31$ \\
OLMo step 1.41M (final) & $\rho = +0.28$, $p = 0.125$, $n = 31$ & $\rho = +0.08$, $p = 0.674$, $n = 31$ \\
Gemma 4 E4B             & $\rho = +0.39$, $p = 0.012$, $n = 41$ & $\rho = -0.17$, $p = 0.299$, $n = 41$ \\
\bottomrule
\end{tabular}
\label{tab:rate-level}
\end{table}

\section{Related Work}
\label{sec:related-work}

\subsection{Spectral and Jacobian analysis of transformers}

\citet{fu2025cast} estimate a population-level linear map $T_i = \tilde{H}_i^\dagger \tilde{H}_{i+1}$ between consecutive layers via least-squares regression, then extract effective rank and spectral decay via SVD. Their compression--expansion cycle in GPT-2 and Llama~3.2~1B resonates with our U-shaped condition number profile, but their $T_i$ is a global regression fit---not a local linearization---and explains only $\sim$62\% of deeper-layer outputs, precisely the regime where the nonlinear dynamics we characterize via Jacobian eigendecomposition are most pronounced.

\citet{golden2025equivalent} construct ``detached Jacobians'' by freezing each nonlinearity at its inference-time value, yielding an exactly linear factorization of the model. SVD of these objects reveals cumulative stable rank collapsing to $\sim$1--3 in Llama~3.2~3B---parallel to our cumulative participation ratio collapse to $\sim$7 in Llama~3.1~8B. Convergence across true and detached Jacobians is strong evidence that progressive dimensional funneling is a genuine structural property of transformer computation. Detached Jacobians cannot characterize perturbation sensitivity or stability, and without eigendecomposition the rotational dynamics encoded in complex eigenvalues remain invisible.

The most closely related empirical work is \citet{aubry2025coupling}, who compute \textit{Residual} Jacobians $\partial f^l / \partial x^{l-1}$ (excluding the skip connection) across 30+ models and document cross-layer singular-vector alignment that correlates with benchmark performance ($R^2 = 0.80$). Our work differs in three respects: (i)~we compute the \textit{full} Jacobian $J_l = I + \partial f_l / \partial h_l$, where the identity term produces the near-unity spectral radius, expanding/contracting mode counts, and non-normality gradient that structure our analysis; (ii)~we perform eigendecomposition, revealing that $\sim$98\% of eigenvalues are complex conjugate pairs encoding rotational dynamics invisible to SVD; (iii)~we separate architecture from training at per-unit granularity via OLMo's step-0 checkpoint, going beyond their existence proof of a training effect to identify \textit{which specific properties} are architectural versus training-induced.

\citet{saratchandran2026spectral} analyze the \textit{parameter} Jacobian $\partial \mathbf{A}/\partial W$ relevant to optimization, reporting condition numbers of $10^9$--$10^{11}$---orders of magnitude larger than our input Jacobian condition numbers ($\kappa \sim 84$--$10^5$), reflecting the dramatic conditioning improvement from the residual connection.

\subsection{Depth as a dynamical system}

The interpretation of network depth as time in a dynamical system originates with \citet{ruthotto2020deep}, who established residual networks as forward Euler discretizations of ODEs and showed that constraining $J_Y F$ to be negative semi-definite or purely imaginary yields provably stable architectures. \citet{storm2024ftle} extracted finite-time Lyapunov exponents from MLP Jacobian products, showing that ridges of large $\lambda_1$ align with decision boundaries and that training self-organizes the FTLE distribution toward zero. \citet{engelken2023gradient} showed that even small Lyapunov spectrum spreads cause exponential ill-conditioning of long-term Jacobians, motivating gradient flossing to compress the spread. These works establish Jacobian products as fundamental diagnostics but are restricted to MLPs or RNNs and operate at initialization or on small-scale networks ($d \leq 256$, $L \leq 16$).

The work most closely related to ours is \citet{jacobs2025block}, who coined ``Dynamical Interpretability'' and applied it to DINOv2-Giant ViTs. They demonstrated directional convergence to angular attractors, self-correcting perturbation dynamics, and collapsing stable rank in late depth---phenomenology consistent with what we observe in LLMs. They linearized the depth flow via Dynamic Mode Decomposition on group-averaged states, obtaining eigenvalues near the positive real axis inside the unit circle. We differ in three respects: (i)~we analyze a decoder-only LLM rather than a ViT, establishing cross-modality generality; (ii)~we compute full per-sample, per-layer Jacobians rather than fitting a single global linear operator to averaged states, revealing the $\sim$98\% complex eigenvalue structure invisible to DMD on averages; (iii)~we provide quantitative per-layer diagnostics (U-shaped condition numbers, participation ratios 9--1860, expanding fraction 33\%--61\%) and a topology$\leftrightarrow$dynamics bridge that separates architectural priors from training-induced properties at per-unit granularity.

\subsection{Residual stream as shared workspace}

\citet{elhage2021framework} introduced the framework of transformer components reading from and writing to a shared residual stream, where each attention head and MLP block projects down, computes, and adds its output back. This additive picture predicts that individual blocks operate as approximately independent perturbations to a shared state. Our Jacobian decomposition $J_\ell = I + R_\ell$ tests this directly: the residual norm ratio $\|R_\ell\|_F / \|J_\ell\|_F$ quantifies how much each block perturbs the residual stream beyond the identity pass-through, and the near-random forward alignment (\S\ref{app:forward-alignment}) confirms that adjacent blocks' leading singular subspaces are largely decoupled---consistent with the shared-workspace model. The one departure is the mid-layer $R$ forward alignment peak in trained OLMo ($3.8\times$ baseline), suggesting that training can install partial sequential coupling between blocks in the depth range where the operator-type transition is steepest.

In the architecture-design direction, \citet{prairie2026parcae} enforce stability in looped transformers by parameterizing the state matrix with guaranteed negative eigenvalues, and \citet{godin2026score} control contraction through explicit ODE step sizes. \citet{koubbi2026homogenized} prove that transformers with independent weights converge to an It\^o SDE on the sphere, predicting representation collapse via logistic dynamics. These theoretical baselines motivate---but do not replace---the direct empirical measurement of trained Jacobian spectra that we provide.

\clearpage

\end{document}